\documentclass[runningheads]{llncs}

\usepackage{eccv}

\usepackage{eccvabbrv}

\usepackage{graphicx}
\usepackage{booktabs}

\usepackage[accsupp]{axessibility}  

\usepackage{amsmath}
\usepackage{amssymb}
\usepackage{subcaption}

\usepackage{hyperref}

\usepackage{orcidlink}

\usepackage[normalem]{ulem}

\usepackage{xspace}
\usepackage{currfile}
\xspaceaddexceptions{]\}}

\newcommand{\nameCOLOR}[1]{\textcolor{black}{#1}} 

\newcommand{\CHORE}{\nameCOLOR{\mbox{CHORE}}\xspace}
\newcommand{\CONTHO}{\nameCOLOR{\mbox{CONTHO}}\xspace}
\newcommand{\hoitg}
{\nameCOLOR{\mbox{HOI-TG}}\xspace}

\newcommand{\nameMethod}{\textcolor{black}{\mbox{MILO}}\xspace}
\newcommand{\nameMethodLONG}{\textbf{M}odeling \textbf{I}nteractions using \textbf{L}arge Reconstruction M\textbf{O}dels}

\newcommand{\lrm}{\nameCOLOR{\mbox{LRM}}\xspace}
\newcommand{\lrmfull}{\nameCOLOR{\mbox{Large Reconstruction Models}}\xspace}

\newcommand{\hoi}{\mbox{{HOI}}\xspace}

\definecolor{myBlue}{rgb}{0.18, 0.38, 0.82}  
\definecolor{myRed}{rgb}{1.0, 0.27, 0.0}
\definecolor{myOrange}{rgb}{0.86, 0.39, 0.12}

\newcommand{\outTitleFULL}{Reconstructing Humans and Objects in Interaction using \lrmfull}

\newcommand{\supmat}{\textcolor{black}{{SuppMat}}\xspace}

\usepackage{lipsum}

\newcommand{\smpl}{\mbox{\nameCOLOR{SMPL}}\xspace}
\newcommand{\smplx}{\mbox{\nameCOLOR{SMPL-X}}\xspace}

\newcommand{\smplh}{\mbox{\nameCOLOR{SMPL-H}}\xspace}

\newcommand{\rgb}{\mbox{RGB}\xspace}

\renewcommand{\etal}{\mbox{et al.}\xspace}
\renewcommand{\ie}{\mbox{i.e.}\xspace}

\newcommand{\twoD}{{2D}\xspace}

\newcommand{\threeD}{\xspace{3D}\xspace}

\newcommand{\GT}{\xspace{GT}\xspace}
\newcommand{\gt}{\GT}

\newcommand{\sota}{{state-of-the-art}\xspace}

\newcommand{\colorRef}[1]{\textcolor{black}{#1}} 
\usepackage[capitalize]{cleveref}
\crefname{figure}{\colorRef{Fig.}}{\colorRef{Figs.}}
\Crefname{figure}{\colorRef{Figure}}{\colorRef{Figures}}
\crefname{section}{\colorRef{Sec.}}{\colorRef{Secs.}}
\Crefname{section}{\colorRef{Section}}{\colorRef{Sections}}
\Crefname{table}{\colorRef{Table}}{\colorRef{Tables}}
\crefname{table}{\colorRef{Tab.}}{\colorRef{Tabs.}}
\Crefname{equation}{\colorRef{Equation}}{\colorRef{Equations}}
\crefname{equation}{\colorRef{Eq.}}{\colorRef{Eqs.}}

\definecolor{lightgray}{gray}{0.97}
\definecolor{lightblue}{rgb}{0.93,0.95,1.0}

\definecolor{GreenColor}{rgb}{0.137,0.573,0.565}
\definecolor{OrangeColor}{rgb}{0.914,0.541,0.0.141}
\definecolor{PurpleColor}{rgb}{0.5,0,0.7}

\newcommand{\cmark}{\color{ForestGreen}\ding{51}}

\newcommand{\xmark}{\color{red}\ding{55}}

\usepackage{multirow}
\usepackage{bbm}

\usepackage{pifont}
\usepackage{xcolor}
\usepackage{wrapfig}
\usepackage{booktabs}

\begin{document}

\title{\outTitleFULL}

\titlerunning{\nameMethod}

\author{Agniv Chatterjee\orcidlink{0000-0001-7199-1196} \and
Georgios Pavlakos\orcidlink{0000-0001-5821-1909}}

\authorrunning{A.~Chatterjee and G.~Pavlakos}

\institute{University of Texas at Austin\\
\email{agniv.chatterjee@utexas.edu, pavlakos@cs.utexas.edu}}

\maketitle

\begin{abstract}

Estimation of Human-Object Interactions in \threeD (\threeD \hoi) is a fundamental problem in \threeD computer vision with applications in AR/VR, robotics, and embodied AI. 
However, reconstructing these interactions in 3D remains challenging due to depth ambiguities, occlusions, and object shape variability.
Existing approaches are primarily concerned with reprojection and contact constraints, fitting parametric human models and object templates to 2D images.
In this paper, we explore a different avenue.
We present \nameMethod, a framework that leverages the visual capabilities of \lrmfull (LRMs) to recover detailed \threeD human-object interactions from a single image.
Our key observation is that LRMs provide a powerful geometric scaffold that preserves relative human-object arrangement and proximity cues.
This significantly simplifies the reconstruction procedure, reframing the problem as interpreting the LRM mesh: we segment it into human and object components, fit a parametric body model to the human part, and optionally align an object template to the object part (if such a template is available).
\nameMethod achieves strong reconstruction accuracy and outperforms existing baselines across multiple benchmarks and interaction scenarios.
Our code is available at \textcolor{blue}{https://ac5113.github.io/MILO}.


\end{abstract}  
\section{Introduction}
\label{sec:intro}

\begin{figure*}
    \vspace{-0.2em}
    \centering
    \includegraphics[width=\textwidth]{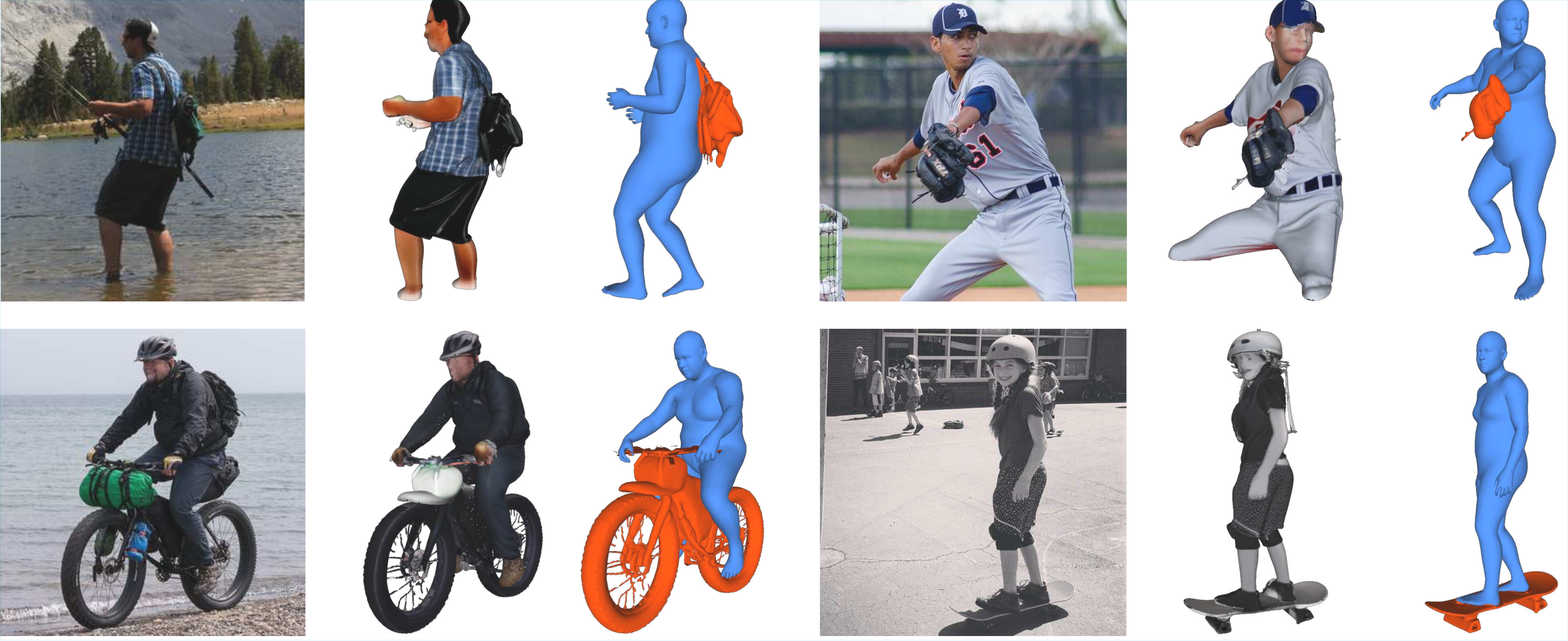}
    \vspace{-1.5em}
    \caption{
    \textbf{We present \nameMethod, an approach for reconstructing \threeD human-object interactions from a single image.} 
        Given an input photo (left), \nameMethod uses a Large Reconstruction Model to predict a combined human-object mesh (center).
        This mesh is interpreted (right) by identifying human and object parts, fitting a \smplh body to the human part, and optionally an object template to the object part.
        The examples illustrate accurate \threeD human mesh together with consistent object pose and shape. 
        Human meshes are shown in \textcolor{myBlue}{blue} and objects in \textcolor{myRed}{red}. 
    }
    \label{fig:teaser}
    \vspace{-2.0em}
\end{figure*}

Daily human life is characterized by rich \threeD interactions with a wide variety of objects, from utensils and smartphones to furniture and tools. 
Modeling these human-object interactions (\hoi) is central to many vision applications, including immersive AR/VR, teleoperation, and robot manipulation. 
However, HOI reconstruction from a single image is difficult due to diverse object shapes and affordances, multiple interaction modes, occlusions, and depth ambiguities, which make the recovery of human pose and object geometry challenging.

Human parametric models such as \smpl~\cite{SMPL:2015}, MANO~\cite{MANO:SIGGRAPHASIA:2017}, and \smplx~\cite{SMPL-X:2019} provide standardized representations which have led to successes for \threeD human reconstruction. 
In contrast, the absence of analogous canonical models for generic objects makes object reconstruction significantly more challenging.
Moreover, the generalization capabilities of methods~\cite{nam2024contho, WangHoiTg} trained on existing \hoi datasets~\cite{bhatnagar2022behave, huang2022intercap} are hindered by the characteristics of these datasets, which are captured in controlled environments with a limited set of objects and scripted interactions. 
Additionally, approaches~\cite{cseke_tripathi_2025_pico, wu2024reconstructing} that require the retrieval of object meshes further depend on the quality and coverage of available shape repositories.
More importantly, accurate placement of retrieved meshes often requires privileged cues such as ground-truth contact or depth~\cite{zhang2020phosa, nam2024contho, cseke_tripathi_2025_pico}. 

Our insight is that we can circumvent many of these limitations by leveraging the rich visual capabilities of \lrmfull (LRMs)~\cite{hunyuan3d22025tencent, lrm, real3d, instantmesh}.
LRMs are trained on large-scale data~\cite{deitke2023objaverse,deitke2023objaverseXL} and are capable of reconstructing complex scenes, outputting an image-specific joint reconstruction of the human and the object.
This effectively provides a geometric scaffold that encodes relative arrangement and proximity cues.
Building on this success, our method proposes to replace the ambiguous image fitting and the contact assumptions with the interpretation of an \lrm-generated human-object reconstruction.

Our goal is to systematically assess the extent to which LRMs can be used for \hoi reconstruction.
While previous work has used LRMs primarily to reconstruct the object~\cite{easyhoi}, our insight is that LRMs can be very powerful for capturing the relative positioning and proximity cues between a human and an object.
To this end, we propose \nameMethod; \nameMethodLONG. 
\nameMethod first applies an off-the-shelf \lrm~\cite{hunyuan3d22025tencent} to the input \rgb image to obtain a holistic human-object mesh.
This reframes our problem as explaining the \lrm interaction mesh.
Through multi-view segmentation of the mesh, we can extract the human and the object point cloud.
For the human part, we fit the parametric \smplh body model to the \lrm mesh via predicted 3D keypoints~\cite{vitpose, hamer}.
For the object, we compose the raw LRM reconstruction with the fitted human, or optionally align a template mesh if one is available.
In that case, alignment is supported by establishing semantic correspondences~\cite{zhang2024telling} between the template and the LRM mesh.
We present example results in \cref{fig:teaser}.

We quantitatively compare \nameMethod with \sota methods on multiple datasets, \ie, InterCap~\cite{huang2022intercap}, HODome~\cite{zhang2023neuraldome} and IMHD~\cite{zhao2024imhoi}, while also showing qualitative results on in-the-wild images~\cite{cseke_tripathi_2025_pico}. 
Importantly, \nameMethod outperforms previous work across the board, without using ground-truth contact information.

Our main contributions are as follows:
\begin{enumerate}
    \item We demonstrate the power of \lrmfull 
    for \hoi reconstruction. Instead of operating with the high ambiguity of reprojection objectives, we reframe the problem as explaining the \lrm mesh results.
    \item We design an approach that robustly fits a parametric human model to the LRM mesh and optionally aligns an object template to it.
    \item We achieve state-of-the-art performance across multiple benchmarks, all while using weaker information than previous work, \ie, without relying on contact information.
\end{enumerate}
\section{Related Works}
\label{sec:related}

\subsection{Monocular \threeD Human Reconstruction}

Single-image \threeD human reconstruction methods can be broadly categorized into parametric and non-parametric approaches. 
Parametric methods~\cite{poco, Kocabas2021pare, li2021hybrik, li2022cliff, rong2021frankmocap} estimate the parameters of a body model~\cite{SMPL:2015, MANO:SIGGRAPHASIA:2017, SMPL-X:2019, xu2020ghum}, whereas non-parametric methods~\cite{Kolotouros2019ConvolutionalMR, meshgraphormer, lin2021end-to-end} predict explicit meshes or point clouds. 
Optimization-based pipelines~\cite{smplify, SMPL-X:2019, xu2020ghum, unitepeople} recover body pose and shape by fitting a parametric body model to \twoD cues, e.g., keypoints, silhouettes, or segmentation, via differentiable rendering. 
In contrast, learning-based approaches directly regress body-model parameters from inputs~\cite{tokenhmr, lin2023osx, li2022cliff, hand4whole, joo2021eft} using deep networks.
While these methods have led to robust monocular human reconstruction, they typically focus on the person in isolation and do not explicitly reason about surrounding objects or human-object contact.
In our method, we use the HMR2.0 network~\cite{goel2023humans} to obtain initial \smpl parameters from the input \rgb image.

\subsection{Monocular \threeD Object Reconstruction}

Monocular \threeD object reconstruction has a long history with a wide range of \threeD representations explored.
Early work employed explicit volumetric~\cite{3dr2n2, imgto3d_voxel1}, point-based~\cite{psgn}, or mesh-based~\cite{psgn, pixel2mesh} representations, while later methods~\cite{deepsdf, nerf} adopted implicit signed-distance or radiance-field representations. 
More recently, \threeD Gaussian splatting~\cite{gaussian_splat} has emerged as an alternative to NeRF-style radiance fields, enabling efficient object reconstruction.
Diffusion-based image-to-\threeD approaches~\cite{realfusion, makeit3d, magic123} and view-conditioned diffusion models~\cite{zero1to3} further improve visual fidelity and handle complex appearance, but 
often rely on per-instance optimization or suffer from multi-view inconsistencies across generated views.

Recent work on \lrmfull~\cite{lrm, lgm, pflrm, triposr, real3d} has enabled efficient and view-consistent reconstruction of diverse object categories. 
By leveraging the \lrm output, \nameMethod obtains an image-specific joint human-object mesh, which captures accurate geometry and proximity information.
For our implementation, we adopt Hunyuan3D-2.0~\cite{hunyuan3d22025tencent}, although we show that our algorithm is not specific to the choice of LRM.

\subsection{Monocular \threeD Human-Object Reconstruction}

Joint reconstruction of humans and objects from a single image has been explored with both learning- and optimization-based approaches.
\hoitg~\cite{WangHoiTg} employs a transformer to jointly reason about pose and interaction.
LEMON~\cite{yang2023lemon} learns \threeD human-object interaction relations from \twoD images by jointly predicting human contact, object affordance, and spatial relations. 
\CONTHO~\cite{nam2024contho} predicts vertex-level contact maps and refines the human-object mesh poses accordingly.
StackFLOW~\cite{stackflow} regresses dense human-object offsets and refines both the human and object reconstructions using stacked normalizing flows.
\CHORE~\cite{xie2022chore} jointly learns a part-correspondence field that links human and object geometry.
TriDi~\cite{tridi} supports multiple conditioning modes by introducing a trilateral diffusion model over humans, objects, and interactions.

A common strategy involves reconstructing the object by retrieving the closest object mesh from existing shape repositories~\cite{deitke2023objaverse}.
PICO~\cite{cseke_tripathi_2025_pico} 
retrieves object candidates,
while Wu et al.~\cite{wu2024reconstructing} uses text-to-\threeD retrieval to obtain an object template.
Related to our work,
LRMs have also been used to reconstruct objects of interaction directly from images, without relying on explicit CAD templates.
For instance, EasyHOI~\cite{easyhoi} employs InstantMesh~\cite{instantmesh} to obtain object meshes in hand-object interaction scenarios.
Our approach also leverages an LRM, but we propose to use these powerful models for joint human-object reconstruction, simplifying the relative positioning of humans and objects. 

Most of the previous methods assume access to object templates and/or \gt contact annotations, which constrains scalability and generalization to novel objects and in-the-wild scenes.
For our approach, the use of an object template is optional, 
while we do not rely on explicit contact information.

\subsection{Human-Object Contact Estimation}

Human-object contact provides a powerful geometric cue that tightly constrains the arrangement of humans and objects. 
Early work modeled contact at a coarse level~\cite{kamakura1980grasp, dillmann2005grasp, hasson2019obman}, over patches~\cite{Fieraru_2021_AAAI, Mueller:CVPR:2021}, or only on the human body surface~\cite{Shimada2020PhysCapTOG, rempe2021humor, Zhang_2021_ICCV}. 
More recent approaches~\cite{RICH, GRAB:2020, zhang2020phosa, Hassan2021posa, tripathi2023deco} estimate dense, vertex-level contact or correspondence maps and exploit them to regularize \hoi reconstruction or scene understanding tasks.
Several \hoi methods integrate contact signals directly into their pipelines. 
\CONTHO~\cite{nam2024contho} uses vertex-level human-object contact maps for mesh refinement. CHORE~\cite{xie2022chore} learns part-to-body correspondence fields that implicitly encode contact patterns. 
PICO~\cite{cseke_tripathi_2025_pico} uses paired human-object contact maps to guide joint human-object fitting.
A related effort, HULC~\cite{shimada2022hulc}, exploits body-scene contact to stabilize monocular motion capture.
In contrast to them, our method does not require contact annotations. 
Instead, we show that we can extract competitive contact estimates from our HOI reconstructions.

\section{Technical Approach}
\label{sec:method}

\begin{figure*}[!t]
    \centering
    \includegraphics[width=\textwidth]{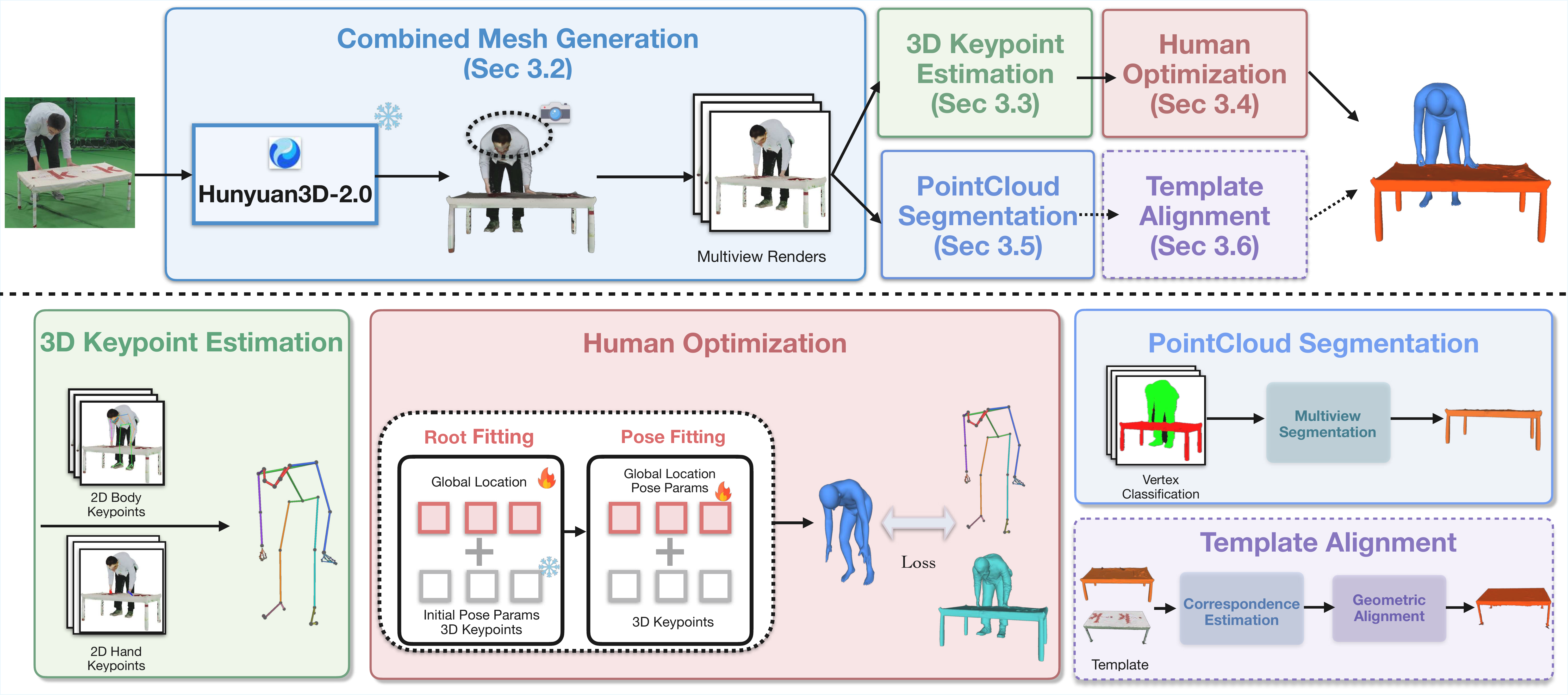}
    \vspace{-1.5em}
    \caption{\textbf{Overview of \nameMethod, our method for reconstructing \threeD human-object interactions from a single image.} 
    Given an input \rgb image, we first use Hunyuan3D-2.0~\cite{hunyuan3d22025tencent} to predict a combined human-object mesh (\cref{subsec:method:hoi}). 
    We then render the mesh from multiple virtual views and estimate \twoD body and hand keypoints in each view (\cref{subsec:method:kp3d}), which are robustly triangulated to obtain \threeD keypoints on the \lrm reconstruction. 
    Starting from an initial \smplh estimate from HMR2.0~\cite{goel2023humans} and HaMeR~\cite{hamer}, we optimize \smplh in two stages (\cref{subsec:method:hum_opt}): (i) \emph{root fitting}, optimizing only global orientation and translation, and (ii) \emph{pose fitting}, additionally optimizing body/hand pose, and shape.
    Finally, we perform multi-view segmentation of the \lrm mesh to extract the object component and compose it with the parametric human mesh for evaluation and visualization (\cref{subsec:method:pc_seg}).
    \emph{If an object template is available}, we optionally align it to the segmented object using semantic correspondences (\cref{subsec:method:temp_align}).}
    \label{fig:full_pipeline}
    \vspace{-1.0em}
\end{figure*}

In this section, we present \nameMethod, our approach that reconstructs detailed human-object interactions from a single \rgb image by leveraging an \lrm.
Sections~\ref{subsec:method: preliminaries} and \ref{subsec:method:hoi} discuss preliminaries, while Sections~\ref{subsec:method:kp3d} and~\ref{subsec:method:hum_opt} describe how we can fit a parametric human model to the \lrm mesh.
In Section~\ref{subsec:method:pc_seg}, we describe how we can segment the reconstructed mesh into a human and object part.
Finally, if an object template is available, Section~\ref{subsec:method:temp_align} describes how we can align it to the \lrm reconstruction.
An overview of our pipeline is shown in \cref{fig:full_pipeline}.

\subsection{Human Body Model Preliminaries}
\label{subsec:method: preliminaries}

We represent the human body with the \smplh model~\cite{MANO:SIGGRAPHASIA:2017}, parameterized by global orientation $\Phi \in \mathbb{R}^3$, body pose $\Theta \in \mathbb{R}^{21\times3}$, hand pose $\Theta_h \in \mathbb{R}^{30\times3}$, shape $\beta \in \mathbb{R}^{10}$, and root translation $\Gamma \in \mathbb{R}^3$.
Given these parameters, \smplh outputs body vertices and kinematic joints as:
\(
    [V, J] = \mathcal{M}(\Phi, \Theta, \Theta_h, \beta, \Gamma)
\label{eq:smpl_verts}
\)
where $V \in \mathbb{R}^{3\times6890}$ are the mesh vertices and $J \in \mathbb{R}^{3\times52}$ are the joint locations.

\subsection{Human-Object Mesh Reconstruction}
\label{subsec:method:hoi}

One of our key ideas is to use an \lrm to directly reconstruct a combined human-object mesh from the input image. 
Concretely, we employ Hunyuan3D-2.0~\cite{hunyuan3d22025tencent}, which produces a combined mesh capturing both the person and the interacted object.
This choice is driven by two key advantages:
(1) LRMs can generate a high-quality \threeD approximation of the object and interaction present in the image.
(2) The combined mesh encodes rich information about human-object spatial relations and proximity cues without requiring explicit contact annotations.
To this end, we pass an RGBA image to the \lrm to obtain a holistic human-object mesh.
The alpha channel represents the combined segmentation maps of the human and the object.
We interpret the \lrm output mesh as a non-parametric interaction scaffold, segment it into human and object components, and fit the \smplh model to the human part to explain how the person interacts with the object.
If an object template is available, we can also align it to the \lrm mesh.

\subsection{\threeD Keypoint Estimation}
\label{subsec:method:kp3d}

The goal of this step is to estimate a sparse set of \threeD keypoints on the \lrm mesh and use them for subsequent \smplh fitting.
Specifically, we render the \lrm mesh from a set of virtual views $\mathcal{V}$ (set to 60) and run ViTPose~\cite{vitpose} and HaMeR~\cite{hamer} on each rendered image to obtain body and hand keypoints, respectively.
For each view $v \in \mathcal{V}$, ViTPose predicts the COCO-25 body keypoints $x_b^{(v)} \in \mathbb{R}^{25 \times 2}$ with confidence scores $c_b^{(v)} \in [0,1]^{25}$. 
Simultaneously, HaMeR predicts hand keypoints $x_{h,s}^{(v)} \in \mathbb{R}^{21 \times 2}$ for $s \in \{\mathrm{L}, \mathrm{R}\}$ (with confidence set to 1).

For each keypoint, we discard low-confidence detections, triangulate the surviving observations across all pairs of observed views, and select the hypothesis with the largest consensus set under a reprojection-error threshold, followed by nonlinear refinement over the inlier views.
This returns the final set of body and hand \threeD keypoints $(X, \bar{c})$, with $X \in \mathbb{R}^{67\times 3}$. The confidence $\bar{c}$ comes from aggregating the per-keypoint confidence over its inlier views.
These keypoints are used in the next stage to fit \smplh to the \lrm mesh. 
We provide details about the confidence thresholds, triangulation, and aggregation in the \supmat.

\subsection{Human Optimization}
\label{subsec:method:hum_opt}

We get initial estimates of the \smplh parameters using HMR2.0~\cite{goel2023humans}, which predicts human pose and shape in the camera frame. 
We also initialize the hand pose parameters using HaMeR~\cite{hamer}.
We then refine these estimates in two stages: the root fitting, which optimizes for the global orientation and the root translation, followed by the pose fitting stage, which additionally optimizes for the body shape, body, and hand pose.
Both stages optimize the \smplh mesh, keeping the \lrm mesh fixed. 
Additionally, in the first fitting stage, we optimize a human scale parameter.
This helps align the human to the \lrm mesh without devolving to degenerate scales, since the \lrm mesh is not guaranteed to produce the combined mesh at the correct metric scale.

\subsubsection{Root Fitting.}
\label{subsubsec:method:root_fit}

Let $\hat{P} = \{\tilde{\Phi}, \tilde{\Theta}, \tilde{\Theta^h}, \tilde{\beta}, \tilde{\Gamma}\}$ denote the initial \smplh prediction for image $I$ as given by HMR2.0~\cite{goel2023humans} and HaMeR~\cite{hamer}, with the parameters expressed in the camera frame.
We initialize the body pose ($\Theta$), hand pose ($\Theta^h$) and body shape ($\beta$) with these values.
In this first stage, we keep $\Theta, \Theta^h, \beta$ fixed and only optimize $\Phi$ and $\Gamma$ to align the \smplh joints $X^S$ to their corresponding 3D keypoints $X$ estimated in the previous section.
The objective is expressed as: 
\(
\mathcal{L}_{rf} = \sum_i \rho (X^S_i - X_i),
\)
where $\rho$ is the Geman-McClure robustifier~\cite{GemanMcclure}.
This stage runs for 30 iterations and coarsely aligns \smplh with the \lrm mesh.

\subsubsection{Pose Fitting.}
\label{subsec:method:pose_fit}

Next, we optimize the full \smplh parameter set, i.e., shape $\beta$, body pose $\Theta$, and hand pose $\Theta^h$.
To keep the solution feasible, we regularize pose and shape with the VPoser prior~\cite{SMPL-X:2019}, the MANO prior~\cite{MANO:SIGGRAPHASIA:2017} and an $L_2$ shape prior.
Let $\zeta_{bp}\in \mathbb{R}^{32}$ be the representation of the body pose parameters $\Theta$ in the latent space of
VPoser~\cite{SMPL-X:2019}, and $\zeta_{hp} \in \mathbb{R}^{30}$ be the representation of the hand pose parameters $\Theta^h$ in the latent space of
the MANO model~\cite{MANO:SIGGRAPHASIA:2017}. 
The regularization terms can be expressed as:
\(
\mathcal{L}_{bp} = \|\zeta_{bp}\|^2, 
\)
\(
\mathcal{L}_{hp} = \|\zeta_{hp}\|^2
\)
and
\(
\mathcal{L}_{\beta} = \|\beta\|^2.
\)

Under severe self-occlusion or object occlusion, the \lrm may hallucinate missing body parts, which can lead to image-inconsistent pose estimates. 
While $\mathcal{L}_{bp}$ and $\mathcal{L}_{hp}$ encourage feasible poses, they do not explicitly preserve consistency with the image. 
To improve robustness in such cases, we add an anchoring loss based on the HMR2.0~\cite{goel2023humans} pose initialization. 
We use the initialized latent body pose
$\zeta_{bp}^{\mathrm{hmr}}$ and penalize large deviations from it during pose fitting:
\begin{equation}
    \mathcal{L}_{ha}
    =
    \sum_j
    \left[
        \big(\zeta_{bp,j} - \zeta_{bp,j}^{\mathrm{hmr}}\big)^2 - \tau
    \right]_+,
\end{equation}
where $\tau$ is a tolerance margin, set to 2.5.
This acts as a soft anchor: small latent refinements are not penalized, while larger deviations are discouraged.

To further improve alignment in occluded regions, we introduce a visible-vertex \threeD consistency loss with the \lrm reconstruction.
Let $P^{\mathrm{obs}}\subset\mathbb{R}^3$ denote a set of \threeD points sampled from the observed (human) \lrm reconstruction, and let $P^{\mathrm{pred}}\subset\mathbb{R}^3$ denote the set of predicted \smplh mesh vertices.
Because the reconstruction can be unreliable in occluded areas, we restrict the comparison to a subset $P^{\mathrm{pred}}_{\mathcal V}\subseteq P^{\mathrm{pred}}$ containing only those predicted vertices associated with visible joints, where $\mathcal V$ is the vertex-visibility set estimated from the rendered views.
We then define a one-way robust Chamfer loss from the observed points to these visible predicted vertices:

\begin{equation}
    \mathcal{L}_{3D}
    =
    \frac{1}{|P^{\mathrm{obs}}|}
    \sum_{p \in P^{\mathrm{obs}}}
    \rho\!\left(
    \min_{q \in P^{\mathrm{pred}}_{\mathcal{V}}}
    \|p-q\|_2
    \right),
\end{equation}
where $p\in P^{\mathrm{obs}}$ is an observed \threeD point, $q\in P^{\mathrm{pred}}_{\mathcal V}$ is a visible predicted vertex, $|P^{\mathrm{obs}}|$ is the number of observed points, and $\rho(\cdot)$ is a robust penalty (Geman--McClure) to reduce sensitivity to outliers.
This term improves fitting in both self-occlusion and object-occlusion cases.

We combine these with the root loss to obtain the final fitting objective:
\begin{equation}
    \mathcal{L}_{pf} =
    \lambda_{rf}\mathcal{L}_{rf}
    + \lambda_{bp}\mathcal{L}_{bp}
    + \lambda_{hp}\mathcal{L}_{hp}
    + \lambda_{\beta}\mathcal{L}_{\beta}
    + \lambda_{ha}\mathcal{L}_{ha}
    + \lambda_{3D}\mathcal{L}_{3D},
\label{eq:smpl_loss}
\end{equation}
where $\lambda_{rf}=10$, $\lambda_{bp}=\lambda_{hp}=0.04$, $\lambda_{\beta}=0.05$, $\lambda_{ha}=10$ and $\lambda_{3D}=50$.
We optimize this objective for 60 iterations, yielding a human mesh that is tightly aligned with the \lrm reconstruction while remaining anatomically plausible.

\subsection{Point Cloud Segmentation}
\label{subsec:method:pc_seg}

The next step is to extract the object geometry from the \lrm mesh to obtain an object point cloud.
We use the same multi-view renders from \cref{subsec:method:kp3d} and apply a segmentation model~\cite{ravi2024sam2segmentimages, liu2023grounding} to obtain human and object masks. 
The prompt used is ``person.<obj$\_$name>'',
indicating the return of one mask for the person and one for the object.
The visible vertices in each view are projected onto the image plane and labeled using these masks, followed by multi-view aggregation and filtering.
For each vertex, we aggregate the per-viewpoint scores that account for both viewpoint quality and a boundary-closeness weight.
The viewpoint quality score balances geometric coverage and segmentation reliability.
The boundary-closeness weight handles human-object contact regions.
Next, we threshold the aggregated scores to obtain final labels.
The vertices labeled as non-human form the raw object point cloud, which we clean via geometric filtering. 
This procedure is described in detail in the \supmat.
The resulting clean object point cloud is composed with the fitted \smplh mesh.

\begin{wrapfigure}{R}{0.5\textwidth}
  \vspace{-3.0em}
  \begin{center}
    \includegraphics[width=0.48\textwidth]{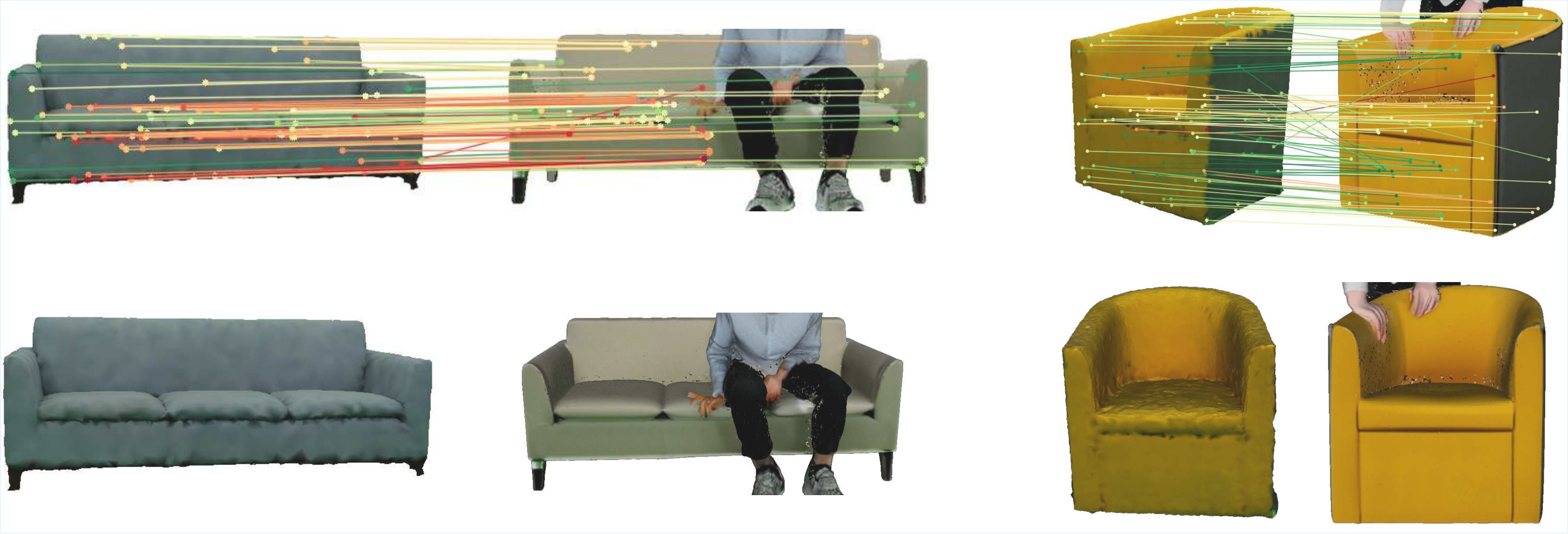}
  \end{center}
  \vspace{-2.0em}
  \caption{\textbf{Template Alignment.} 
  Top row: Visualization of semantic correspondences between a render of the template mesh and the \lrm mesh.
  Bottom row: Alignment of the template mesh with the \lrm mesh.
  }
  \label{fig:temp_align}
  \vspace{-1.5em}
\end{wrapfigure}

\subsection{Template Alignment}
\label{subsec:method:temp_align}

\nameMethod can operate without a template at test time. 
When a dataset provides a template mesh, we can optionally align it to the segmented object component of the \lrm reconstruction. 
Let $M^{\mathrm{tmp}}=\{u_i\}_{i=1}^{N}$ be the template mesh vertices and let $P^{\mathrm{lrm}}=\{v_j\}_{j=1}^{M}$ denote the segmented \lrm point cloud. 
We seek a similarity transform:
\(
T(u)=s\mathbf{R}u+\mathbf{t},
\) with
\(
s>0,
\)
\(
\mathbf{R}\in SO(3),
\)
and
\(
\mathbf{t}\in\mathbb{R}^3,
\)
aligning the template to the reconstructed object.

\noindent
\textit{Semantic correspondences.}
Our first step is to establish correspondences between the LRM and the template mesh.
We do this by considering multi-view renders of both meshes.
For this, we first restrict matching to \lrm renders that are most consistent with the input image.
Using the geometry-aware semantic correspondence model of Zhang~\etal~\cite{zhang2024telling}, we compute a global mutual-nearest-neighbor similarity \(S(\cdot,\cdot)\) between the input image and every \lrm render (considering only the object foreground), and retain the top-\(K_{\mathrm{in}}\) views with \(K_{\mathrm{in}}=5\).

Next, for each template render \(I_q^{\mathrm{tmp}}\), we evaluate its similarity \(S(I_q^{\mathrm{tmp}}, I_r^{\mathrm{lrm}})\) against the retained \lrm renders $I_r^{\mathrm{lrm}}$ and keep the top-\(K_{\mathrm{pair}}\) matches (\(K_{\mathrm{pair}}=3\)). 
For each selected pair \((q,r)\), we compute dense bidirectional pixel correspondences in descriptor space. Let \(\Pi_{q\rightarrow r}(x)\) denote the target pixel matched to source pixel \(x\), and let \(C_{q\rightarrow r}(x)\) be its cosine similarity.

We lift these \twoD matches to \threeD correspondences by unprojection. 
A template vertex \(u_i\) is considered only if it is visible in view \(q\), projects inside the foreground mask, and its matched pixel \(\Pi_{q\rightarrow r}(\pi_q(u_i))\) lies within a radius \(\tau_{2\mathrm{D}}\) of some visible \lrm vertex projection in view \(r\). 
The matched \lrm vertex is:
\begin{equation}
j^*(i,q,r)=
\arg\min_{j\in\mathcal{V}_r^{\mathrm{lrm}}}
\left\|
\pi_r(v_j)-\Pi_{q\rightarrow r}(\pi_q(u_i))
\right\|_2,
\end{equation}
subject to:
\(
C_{q\rightarrow r}(\pi_q(u_i))>\tau_{\mathrm{loc}}
\)
and
\(
\left\|
\pi_r(v_{j^*})-\Pi_{q\rightarrow r}(\pi_q(u_i))
\right\|_2<\tau_{2\mathrm{D}}.
\)
We use \(\tau_{\mathrm{loc}}=0.6\) and \(\tau_{2\mathrm{D}}=15\) pixels.
A template vertex may obtain multiple candidate matches across view pairs. 
We retain only the highest-scoring one, with score:
\begin{equation}
\omega_{i,q,r}
=
C_{q\rightarrow r}(\pi_q(u_i))\cdot S(I_q^{\mathrm{tmp}}, I_r^{\mathrm{lrm}}).
\end{equation}
This gives a sparse forward map \(i\mapsto j_i\) with weight \(w_i=\max_{q,r}\omega_{i,q,r}\).
To suppress unstable matches, we also compute the reverse map \(\lrm\ \rightarrow\ \)template and apply neighborhood-based cycle consistency. 
A forward match \(u_i\mapsto v_{j_i}\) is retained only if the reverse map sends \(v_{j_i}\) back to a template vertex within a local neighborhood of \(u_i\), defined relative to the local template mesh scale. 
This yields the final sparse correspondence set:
\(
\mathcal{C}=\{(i_k,j_k,w_k)\}_{k=1}^{K},
\)
(\cref{fig:temp_align}).

\noindent
\textit{Template alignment.}
Given sparse weighted correspondences $\mathcal{C}=\{(i_k,j_k,w_k)\}_{k=1}^{K}$, we first estimate an initial similarity transform $T(u)=s\mathbf{R}u+\mathbf{t}$ by solving a weighted Sim(3) Procrustes problem with a weighted Kabsch fit.
We then refine $T$ with an iterative weighted Sim(3) re-estimation that jointly uses $\mathcal{C}$ and nearest-neighbor constraints to the segmented \lrm object point cloud $P^{\mathrm{lrm}}$.

At iteration $t$, we transform template vertices as $\tilde{u}_i^{(t)}=T^{(t)}(u_i)$, assign each $\tilde{u}_i^{(t)}$ its closest point $p_i^{(t)}=\operatorname{NN}(\tilde{u}_i^{(t)},P^{\mathrm{lrm}})$, and form an inlier set $\mathcal{V}_t=\{i \mid \|\tilde{u}_i^{(t)}-p_i^{(t)}\|_2 < \tau_t\}$ using a threshold $\tau_t$ that is progressively tightened across iterations.
We then compute $T^{(t+1)}$ by a \emph{single} weighted Kabsch-with-scale fit on the combined pairs
$\{(u_{i_k},v_{j_k})\}_{k=1}^{K} \cup \{(u_i,p_i^{(t)})\}_{i\in\mathcal{V}_t}$,
using weights $w_k$ for semantic pairs and a constant weight of $10$ for each nearest-neighbor pair.
This procedure retains the semantic alignment captured by $\mathcal{C}$ while improving surface-level agreement with the reconstructed object geometry.

\section{Experiments}
\label{sec:experiment}

This section presents our empirical evaluation. 
\Cref{subsec:indoor_hoi} compares \nameMethod with the state-of-the-art on the InterCap~\cite{huang2022intercap}, HODome~\cite{zhang2023neuraldome} and IMHD~\cite{zhao2024imhoi} datasets.
We evaluate settings with and without access to a template.
\cref{subsec:contact_est} evaluates the effect of \nameMethod on the downstream task of estimating vertex-level contact on the human body.
\Cref{subsec:scaffold} studies the effect of the underlying \lrm mesh geometry and the dependence of \nameMethod on the choice of \lrm.
\Cref{subsec:easyhoi_exp} compares our joint reconstruction scaffold against object-only and individual-reconstruction baselines. 
\Cref{subsec:ablation} analyzes and ablates the key design choices.
Finally, \cref{subsec:itw_hoi} presents qualitative in-the-wild reconstructions on PICO-db~\cite{cseke_tripathi_2025_pico}.

\begin{table}[t]
\centering
\resizebox{0.8\columnwidth}{!}{
\begin{tabular}{lccc|ccc}
\hline
\multirow{2}{*}{Methods} & \multirow{2}{*}{Type} & \multirow{2}{*}{\begin{tabular}{c}
Needs\\Template?\end{tabular}}& \multirow{2}{*}{\begin{tabular}{c}
Needs\\Contact?\end{tabular}} & PA-CD$_h$ & PA-CD$_o$ & PA-CD$_{h+o}$ \\
 & & & & (cm) $\downarrow$ & (cm) $\downarrow$ & (cm) $\downarrow$ \\ 
\hline
CONTHO~\cite{nam2024contho} & Reg. & \cmark & \cmark & 8.36 & 24.30 & 13.14 \\
HOI-TG~\cite{WangHoiTg} & Reg. & \cmark & \xmark & 8.22 & 25.05 & 14.63 \\ 
\hline
PHOSA~\cite{zhang2020phosa} & Opt. & \cmark & \cmark & 10.07 & 23.36 & 13.38 \\
Open3DHOI~\cite{wen2025reconstructing} & Opt. & \xmark & \xmark & \underline{6.88} & 31.18 & 10.17 \\
PICO~\cite{cseke_tripathi_2025_pico} & Opt. & \cmark & \cmark & 7.43 & 21.85 & 10.33 \\ 
\hline
Ours \scriptsize{(w/ template)} & Opt. & \cmark & \xmark & 6.96 & \bf 18.97 & \bf 7.45 \\ 
Ours \scriptsize{(w/o template)} & Opt. & \xmark & \xmark & \bf 6.85 & \underline{20.74} & \underline{9.36} \\
\hline
\end{tabular}
}
\caption{\textbf{Quantitative comparison with the state of the art on the InterCap dataset~\cite{huang2022intercap}.} 
We report Procrustes-Aligned Chamfer Distance (PA-CD) for the human mesh (PA-CD$_h$), object mesh (PA-CD$_o$), and the combined human-object pair (PA-CD$_{h+o}$); lower is better.
Regression (Reg.) and optimization (Opt.) baselines are evaluated under the training protocol from Cseke \etal~\cite{cseke_tripathi_2025_pico}. Our method, \nameMethod, uses no contact information, yet achieves the best overall reconstruction accuracy.}
\vspace{-2.0em}
\label{tab:quant_results_icap}
\end{table}

\begin{center}
\begin{minipage}[t]{0.48\columnwidth}
\centering
\resizebox{\linewidth}{!}{
\begin{tabular}{l|ccc}
\hline
Methods & PA-CD$_h$ & PA-CD$_o$ & PA-CD$_{h+o}$ \\
 & (cm) $\downarrow$ & (cm) $\downarrow$ & (cm) $\downarrow$ \\
\hline
PICO~\cite{cseke_tripathi_2025_pico} & 13.12 & 14.44 & 10.07 \\
Ours \scriptsize{(w/ template)} & \bf 9.50 & \bf 12.97 & \underline{6.68} \\
Ours \scriptsize{(w/o template)} & \underline{9.71} & \underline{13.75} & \bf 6.38 \\
\hline
\end{tabular}
}
\captionof{table}{\textbf{Quantitative evaluation on the HODome dataset~\cite{zhang2023neuraldome}.}}
\label{tab:hodome_results}
\end{minipage}\hfill
\begin{minipage}[t]{0.48\columnwidth}
\centering
\resizebox{\linewidth}{!}{
\begin{tabular}{l|ccc}
\hline
Methods & PA-CD$_h$ & PA-CD$_o$ & PA-CD$_{h+o}$ \\
 & (cm) $\downarrow$ & (cm) $\downarrow$ & (cm) $\downarrow$ \\
\hline
PICO~\cite{cseke_tripathi_2025_pico} & 15.81 & 17.70 & 13.24 \\
Ours \scriptsize{(w/ template)} & \underline{11.76} & \underline{13.39} & \underline{10.10} \\
Ours \scriptsize{(w/o template)} & \bf 8.99 & \bf 9.71 & \bf 6.98 \\
\hline
\end{tabular}
}
\captionof{table}{\textbf{Quantitative evaluation on the IMHD dataset~\cite{zhao2024imhoi}.}}
\label{tab:imhd_results}
\end{minipage}
\end{center}

\subsection{Comparison with the state-of-the-art}
\label{subsec:indoor_hoi}
InterCap~\cite{huang2022intercap} is an indoor dataset with ground-truth human and object meshes, including ten human subjects and ten everyday objects in various interaction scenarios. 
We evaluate \nameMethod for \hoi reconstruction in this controlled setting and compare it with both regression-based~\cite{nam2024contho, WangHoiTg} and optimization-based methods~\cite{zhang2020phosa, cseke_tripathi_2025_pico, wen2025reconstructing} following the protocol of Cseke et al.~\cite{cseke_tripathi_2025_pico}. 
For the learning-based baselines, we use checkpoints trained only on BEHAVE~\cite{bhatnagar2022behave}, i.e., enforcing an out-of-domain evaluation.

We report Procrustes-Aligned Chamfer Distance (PA-CD) for the human mesh, the object mesh, and the combined human-object pair. 
The rigid alignment is performed on the combined mesh; after alignment, PA-CD is computed separately for the human and object components as well as for the joint reconstruction.
Quantitative results are presented in \cref{tab:quant_results_icap}. 
\nameMethod outperforms all previous work (even without access to a template), while relying on less privileged information than competing methods.

\begin{figure*}[!t]
    \vspace{-0.1 em}
    \centering
    \includegraphics[width=\textwidth]{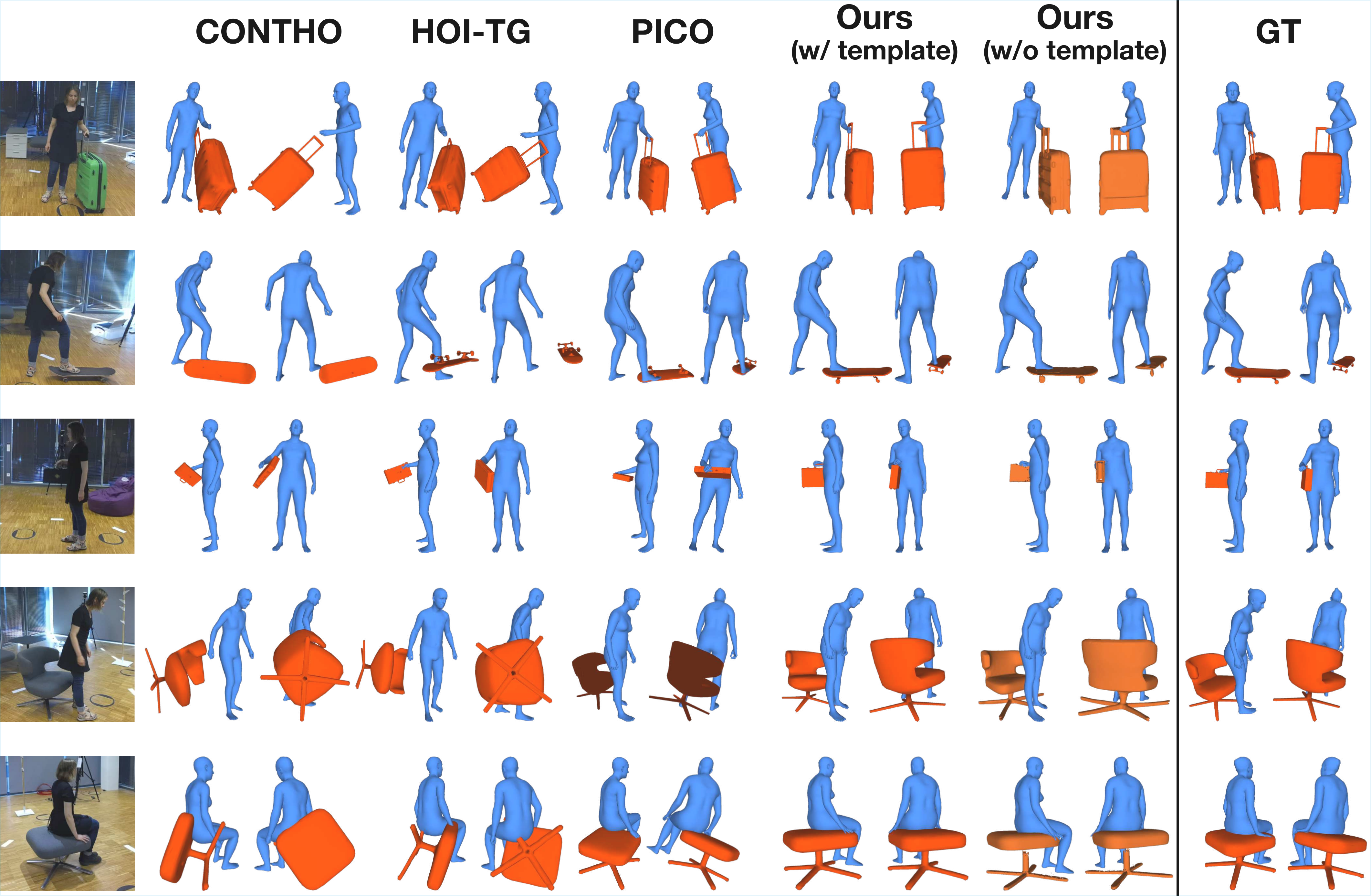}
    \vspace{-2.0em}
    \caption{\textbf{Qualitative comparison on the InterCap dataset~\cite{huang2022intercap}.} 
    From left to right: input \rgb image, CONTHO~\cite{nam2024contho}, HOI-TG~\cite{WangHoiTg}, PICO-fit~\cite{cseke_tripathi_2025_pico}, \nameMethod reconstructions (with and without template meshes, respectively), and ground-truth human-object meshes. 
    Our method achieves better human-object alignment in \threeD.
    Humans are rendered in \textcolor{myBlue}{blue}, template objects in \textcolor{myRed}{red}, and segmented object meshes in \textcolor{myOrange}{orange}.}
    \label{fig:qual_results}
\end{figure*}

\begin{figure*}[!ht]
    \vspace{-0.1 em}
    \centering
    \includegraphics
    [trim=6.0em 0em 0em 0em, clip, width=\textwidth]{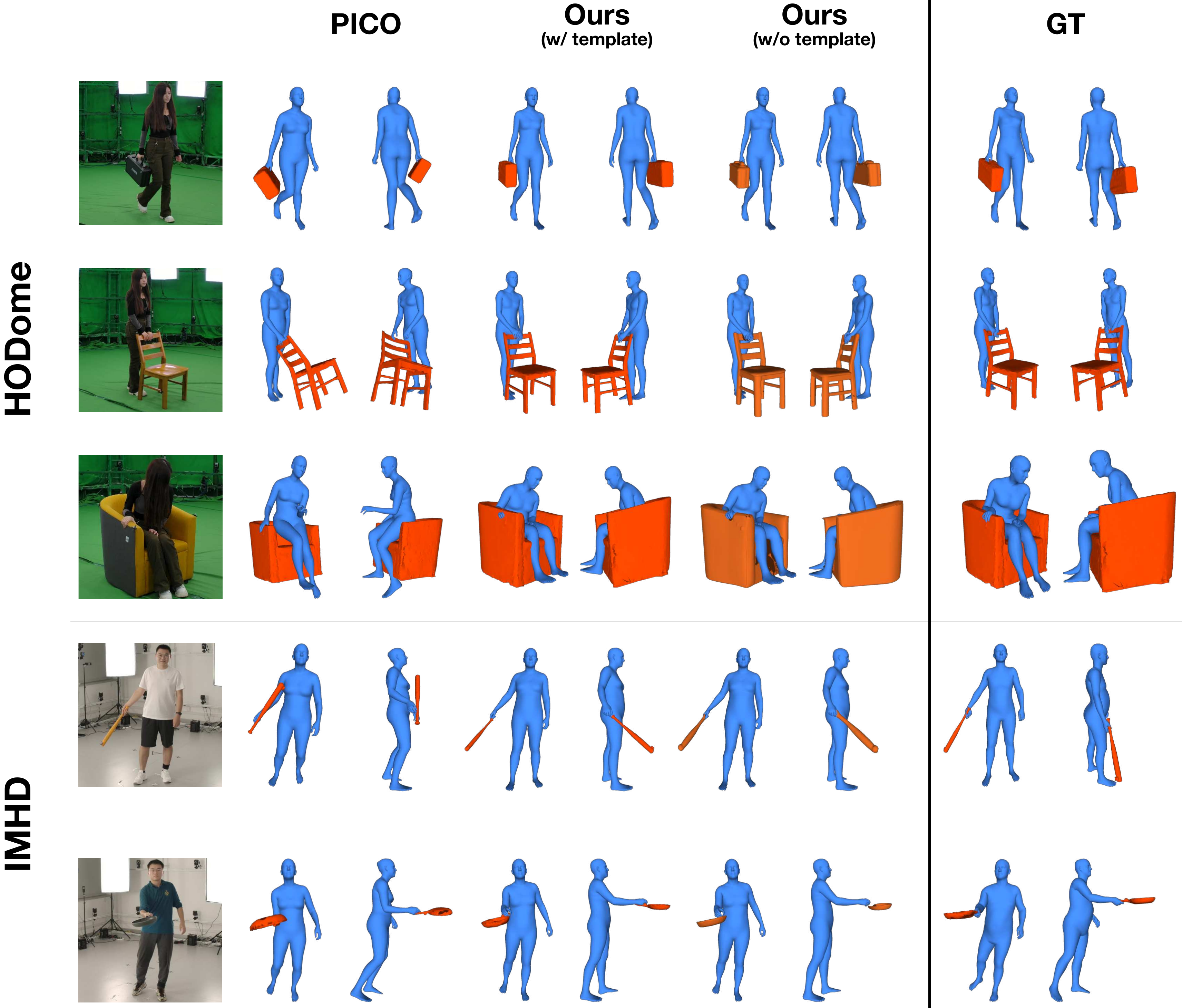}
    \vspace{-2.0em}
    \caption{\textbf{Qualitative evaluation of PICO-fit~\cite{cseke_tripathi_2025_pico} and \nameMethod on images from the HODome~\cite{zhang2023neuraldome} and IMHD~\cite{zhao2024imhoi} datasets.} 
    Our outputs, with or without templates, are more physically plausible and closer to the GT than PICO.
    Humans are rendered in \textcolor{myBlue}{blue}, template objects in \textcolor{myRed}{red}, and segmented object meshes in \textcolor{myOrange}{orange}.}
    \label{fig:qual_results_hodome_imhd}
    \vspace{-0.5em}
\end{figure*}

Additionally, we evaluate \nameMethod on HODome~\cite{zhang2023neuraldome} and IMHD~\cite{zhao2024imhoi} datasets, comparing against the best-performing baseline PICO.
HODome and IMHD are also indoor datasets with ground-truth human and object meshes.
HODome consists of 10 human subjects and 20 object classes, while IMHD consists of 8 human subjects and 10 object categories with different interactions with each object.
As can be seen from \cref{tab:hodome_results} and \cref{tab:imhd_results}, \nameMethod performs better than PICO across the board, in both modes.

The improvements from \nameMethod are also evident in the qualitative results on InterCap~\cite{huang2022intercap} (\cref{fig:qual_results}), HODome~\cite{zhang2023neuraldome}, and IMHD~\cite{zhao2024imhoi} (\cref{fig:qual_results_hodome_imhd}), where \nameMethod achieves better \threeD alignment between the human and the object, with more accurate contact and fewer penetrations than prior approaches.

\subsection{Contact estimation}
\label{subsec:contact_est}

\begin{table}[t]
    \centering
    \scriptsize
    \setlength{\tabcolsep}{5pt}
    \resizebox{0.75 \columnwidth}{!}{
    \begin{tabular}{c|cccc}
        Method & F1$\uparrow$ & Precision$\uparrow$ & Recall$\uparrow$ & geo. (cm)$\downarrow$ \\ 
        \hline
        DECO~\cite{tripathi2023deco} & 9.37 & 12.57 & 10.28 & 126.70 \\
        Ours & \bf 30.00 & \bf 36.57 & \bf 39.49 & \bf 39.57 \\ 
    \end{tabular}
    }
    \caption{\textbf{Contact evaluation.} Metrics are reported on InterCap. Our 3D geometry-inferred contact outperforms the image-regressed contact by the DECO baseline~\cite{tripathi2023deco}.}
    \label{table:contact_comp}
    \vspace{-3.0 em}
\end{table}

\nameMethod does not predict contact, but we can infer contact based on the proximity of the \smplh mesh and the aligned object template. 
We report the results of this strategy in \cref{table:contact_comp}.
Compared to the DECO RICH checkpoint, \nameMethod achieves higher F1, Precision, and Recall, and lower geometric error on InterCap.

\begin{table}[t]
\centering\small
\setlength{\tabcolsep}{4.5pt}
\begin{tabular}{@{}lccc@{}}
\toprule
Mesh Geometry & PA-CD$_h$ (cm) $\downarrow$ & PA-CD$_o$ (cm) $\downarrow$ & PA-CD$_{h+o}$ (cm) $\downarrow$ \\
\midrule
InstantMesh~\cite{instantmesh} & 7.51 & 33.21 & 10.84 \\
SAM3D (joint)~\cite{sam3dteam2025sam3d3dfyimages} & 7.00 & 21.99 & 9.80  \\
Hunyuan3D-2.0~\cite{hunyuan3d22025tencent} & \textbf{6.85} & \textbf{20.74} & \textbf{9.36} \\
\midrule
Oracle & 3.28  & 4.93  & 3.12  \\
\bottomrule
\end{tabular}
\caption{\textbf{Effect of mesh geometry.} Results presented on InterCap.
We replace the mesh reconstructed by Hunyuan3D-2.0~\cite{hunyuan3d22025tencent} with meshes produced by SAM3D~\cite{sam3dteam2025sam3d3dfyimages} and 
InstantMesh~\cite{instantmesh}, as well as the ground-truth geometry (oracle).
We observe that SAM3D performs comparably to Hunyuan3D-2.0, which indicates that \nameMethod is not tied to a specific \lrm choice.
}
\label{tab:scaffold_ablation}
\end{table}

\subsection{Effect of mesh geometry}
\label{subsec:scaffold}

Next, we evaluate the effect of the initial mesh geometry in the \nameMethod pipeline.
Table~\ref{tab:scaffold_ablation} compares our default Hunyuan3D-2.0 choice with alternative \lrm backbones, and an oracle mesh obtained from the ground-truth human-object mesh.
Replacing Hunyuan3D-2.0~\cite{hunyuan3d22025tencent} with SAM3D (joint)~\cite{sam3dteam2025sam3d3dfyimages} gives comparable performance, suggesting that \nameMethod is not tied to a specific \lrm choice. 
InstantMesh~\cite{instantmesh} performs worse, but we acknowledge that it corresponds to an older method.
Finally, the oracle result indicates that the remaining error is largely bounded by upstream reconstruction quality.

\subsection{Comparison with individual reconstruction}
\label{subsec:easyhoi_exp}

\begin{table}[t]
\centering
\resizebox{0.9\columnwidth}{!}{%
\begin{tabular}{c|c|ccc}
\hline
Baseline & Contact Loss & PA-CD$_h$ (cm) $\downarrow$ & PA-CD$_o$ (cm) $\downarrow$ & PA-CD$_{h+o}$ (cm) $\downarrow$ \\
\hline
 EasyHOI* (full-body) & \xmark & 41.57 & 77.06 & 37.21 \\
 EasyHOI* (full-body) & \cmark & 9.48 & 27.93 & 12.08 \\
 \hline
 SAM3D (independent) & \xmark & 18.18 & 31.07 & 15.89 \\
 \hline
 Ours & \xmark & \bf 6.85 & \bf 20.74 & \bf 9.36 \\
\hline
\end{tabular}
}
\caption{\textbf{Effect of joint human \& object reconstruction scaffold.}
Results are presented on InterCap.
The two EasyHOI-style variants follow an object-only pipeline: the object is reconstructed by the \lrm, and its pose is optimized by silhouette reprojection, either alone (first row) or with an additional contact term (second row).
SAM3D (independent) instead reconstructs the human and object independently and composes them through MoGe~\cite{moge} pointmap alignment.
The use of a joint human-object scaffold proposed by \nameMethod (last row) outperforms all baselines.
}
\label{tab:ablation_easyhoi}
\end{table}

\nameMethod leverages the \lrm reconstruction as a shared geometric scaffold for \hoi reconstruction: it fits a parametric human model to the human region of the reconstructed mesh and extracts the interacting object from the same scaffold. 
In contrast, other recent methods~\cite{easyhoi} follow an object-only pipeline, in which an \lrm is used to reconstruct the object mesh and the resulting shape is subsequently fitted using reprojection and contact constraints. 
For example, EasyHOI~\cite{easyhoi} adopts this strategy for hand-object interaction reconstruction.

Although EasyHOI is not directly comparable to our setting because it considers hand-object rather than full-body interactions, we design a baseline that follows the same object-only paradigm. 
Specifically, we follow the same pipeline with EasyHOI~\cite{easyhoi}; first (1) we perform inpainting, then (2) we reconstruct the object with InstantMesh~\cite{instantmesh} and post-process the output mesh to make it watertight, and finally (3) we optimize the object pose using a silhouette reprojection loss.
Additionally, we consider a version where a contact loss is added.
We discuss details of contact estimation in the \supmat.

As shown in \cref{tab:ablation_easyhoi}, \nameMethod outperforms both versions, with and without contact loss.
This result indicates that using a joint reconstruction scaffold is more effective for \hoi reconstruction than an object-only reconstruction followed by fitting to the image. Moreover, we observe that contact cues are essential for these image-fitting approaches, since otherwise the location and scale of the object are mostly unconstrained (first row of \cref{tab:ablation_easyhoi}). Removing the reliance on explicit contact estimation is one of the key advantages of \nameMethod.

To further identify whether MILO's gains come from the joint reconstruction or from the downstream fitting pipeline, we add a controlled baseline that reconstructs the human and the object separately.
We follow the composition method of SAM3D~\cite{sam3dteam2025sam3d3dfyimages, sam3dbody} and run SAM3D independently on the human and the object regions of the same image.
Then, we align the two resulting meshes in a common frame via MoGe~\cite{moge} pointmap alignment, yielding the ``SAM3D (independent)'' baseline in \cref{tab:ablation_easyhoi}.
This baseline differs from the ``SAM3D (joint)'' baseline of Table~\ref{tab:scaffold_ablation} in that the geometry comes from two independent reconstructions instead of a single joint scaffold, so we can isolate the value of joint reconstruction.
As reported in \cref{tab:ablation_easyhoi}, \nameMethod clearly outperforms this baseline across all metrics.
This substantial improvement confirms that the joint scaffold, rather than the fitting procedure, is the main driver of MILO's accuracy.

\subsection{Ablation Experiments}
\label{subsec:ablation}

We also evaluate the significance of the two fitting stages of the optimization pipeline. 
We perform this ablation using both the segmented \lrm mesh and the aligned template (\cref{tab:ablation_stages}).
As can be seen, removing the pose fitting or both fitting stages consistently degrades performance, confirming that both stages of our optimization are necessary to achieve the best overall reconstruction quality.
Additional ablation experiments have been included in the \supmat.

\begin{table}[t]
\centering
\resizebox{1.0\columnwidth}{!}{%
\begin{tabular}{c|c|ccc||c|c|ccc}
\hline
\multirow{2}{*}{\begin{tabular}{c}Human\\Fitting\end{tabular}} & \multirow{2}{*}{\begin{tabular}{c}Uses\\Template\end{tabular}} & PA-CD$_h$ & PA-CD$_o$ & PA-CD$_{h+o}$ & \multirow{2}{*}{\begin{tabular}{c}Human\\Fitting\end{tabular}} & \multirow{2}{*}{\begin{tabular}{c}Uses\\Template\end{tabular}} & PA-CD$_h$ & PA-CD$_o$ & PA-CD$_{h+o}$ \\
& & (cm) $\downarrow$ & (cm) $\downarrow$ & (cm) $\downarrow$ & & & (cm) $\downarrow$ & (cm) $\downarrow$ & (cm) $\downarrow$ \\
\hline
 No fitting & \xmark & 30.58 & 23.53 & 14.10 &  No fitting & \cmark & 14.97 & 37.02 & 14.57 \\
 Root Fitting & \xmark & 7.98 & 22.53 & 10.15 & Root Fitting & \cmark & 7.79 & \bf 18.63 & 7.99 \\
 Pose Fitting & \xmark & \bf 6.85 & \bf 20.74 & \bf 9.36 & Pose Fitting & \cmark & \bf 6.96 & 18.97 & \bf 7.45 \\
\hline
\end{tabular}
}
\caption{\textbf{Effect of human fitting.} Results are presented on the InterCap dataset~\cite{huang2022intercap}. \textbf{No fitting} computes the CD metrics using the raw \lrm mesh (adopting our point-cloud segmentation), without \smplh optimization. \textbf{Root fitting} optimizes only the global orientation and translation of the human. \textbf{Pose fitting} additionally optimizes body/hand pose and shape for SMPL-H.
We consider both template and template-free settings.
The progressive reductions across metrics highlight the importance of the fitting stages and the necessity of a parametric model for precise \threeD \hoi reconstruction.}
\label{tab:ablation_stages}
\end{table}

\begin{figure*}[t]
    \vspace{-0.2 em}
    \centering
    \includegraphics[width=\linewidth]{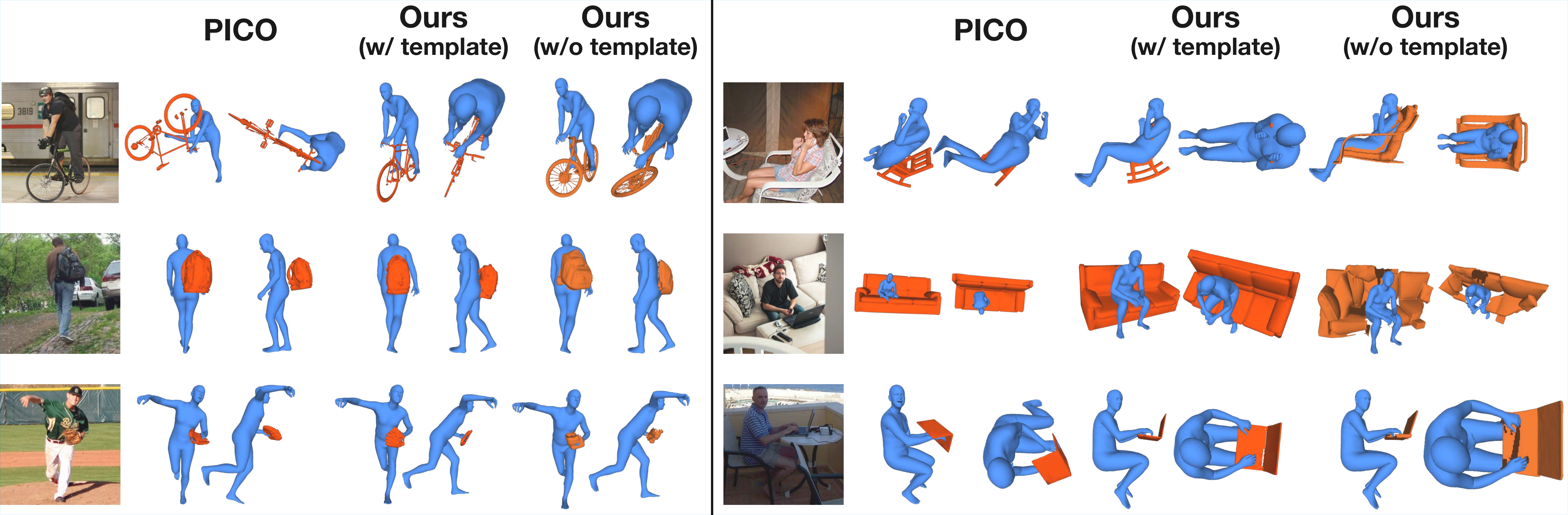}
    \vspace{-1.5em}
    \caption{\textbf{Qualitative reconstructions on in-the-wild images from PICO-db~\cite{cseke_tripathi_2025_pico}.} 
    \nameMethod recovers 
    \threeD human-object geometry from a single image, with coherent arrangement and fewer interpenetrations than the PICO-fit~\cite{cseke_tripathi_2025_pico} baseline. 
    Humans are rendered in \textcolor{myBlue}{blue}, template objects in \textcolor{myRed}{red}, and the segmented object mesh in \textcolor{myOrange}{orange}.}
    \label{fig:qual_results_itw}
    \vspace{-1.0em}
\end{figure*}

\subsection{Qualitative Evaluation on In-The-Wild Images}
\label{subsec:itw_hoi}

To assess \nameMethod's robustness, we further evaluate it qualitatively on in-the-wild images from PICO-db~\cite{cseke_tripathi_2025_pico}.
As illustrated in \cref{fig:qual_results_itw}, \nameMethod recovers human and object meshes that are better aligned in \threeD and 
exhibit more coherent contact patterns and relative positioning compared to prior work~\cite{cseke_tripathi_2025_pico}.

\section{Conclusion}
\label{sec:conclusion}

We introduced \nameMethod, an approach for reconstructing human-object interactions from a single image. 
\nameMethod leverages the visual capabilities of Large Reconstruction Models to predict a holistic human-object mesh.
It then segments the mesh into human and object parts, fits the \smplh body to the human part, and optionally the object template to the object part, if such a template is available.
By treating the \lrm output as a geometric scaffold, our method captures arrangement and proximity cues while remaining agnostic to object category and shape.
Quantitative experiments on multiple benchmarks show that \nameMethod surpasses existing \hoi reconstruction methods despite using significantly less privileged information. 
Moreover, qualitative results on in-the-wild images demonstrate the generalization to diverse objects, poses, and interaction contexts. 
These results highlight the promise of LRMs for scalable, label-efficient \hoi estimation.

\noindent
\textbf{Limitations and Future Work.} 
One limitation of our approach is that the performance of \nameMethod 
is bounded by the quality of the underlying \lrm.
However, given the quick progress in this space~\cite{triposr, sv3d, sam3dteam2025sam3d3dfyimages}, we expect that the performance of LRMs will continue to improve.
Based on our results, we also observe that our performance is hindered by the accuracy of the point cloud segmentation.
We anticipate that with better methods for point-cloud segmentation, our approach will also benefit.
Future work includes the refinement of \lrm outputs, especially for smaller objects, integrating stronger instance-level reasoning and segmentation to extract a better object mesh from the \lrm output, and extending the framework to video and multi-person/multi-object interaction settings.

\section*{Acknowledgement}
\label{sec:acknowledgement}
This research was supported by NSF-2504906, and 2544200; gifts from Adobe, Google, and Nvidia; and computing support on the Vista GPU Cluster through the Center for Generative AI (CGAI) and the Texas Advanced Computing Center (TACC) at the University of Texas at Austin.


%
%
\bibliographystyle{splncs04}
\bibliography{main}

\begin{thebibliography}{10}
\providecommand{\url}[1]{\texttt{#1}}
\providecommand{\urlprefix}{URL }
\providecommand{\doi}[1]{https://doi.org/#1}

\bibitem{dillmann2005grasp}
Bernardin, K., Ogawara, K., Ikeuchi, K., Dillmann, R.: A sensor fusion approach for recognizing continuous human grasping sequences using hidden markov models. {Transactions on Robotics (T-RO)}  \textbf{21}(1),  47--57 (2005)

\bibitem{bhat2023zoedepth}
Bhat, S.F., Birkl, R., Wofk, D., Wonka, P., M{\"u}ller, M.: Zoedepth: Zero-shot transfer by combining relative and metric depth. arXiv preprint arXiv:2302.12288  (2023)

\bibitem{bhatnagar2022behave}
Bhatnagar, B.L., Xie, X., Petrov, I.A., Sminchisescu, C., Theobalt, C., Pons-Moll, G.: {BEHAVE}: {D}ataset and method for tracking human object interactions. In: {Computer Vision and Pattern Recognition (CVPR)}. pp. 15935--15946 (2022)

\bibitem{smplify}
Bogo, F., Kanazawa, A., Lassner, C., Gehler, P., Romero, J., Black, M.J.: Keep it {SMPL}: {A}utomatic estimation of {3D} human pose and shape from a single image. In: {European Conference on Computer Vision (ECCV)}. vol.~9909, pp. 561--578 (2016)

\bibitem{3dr2n2}
Choy, C.B., Xu, D., Gwak, J., Chen, K., Savarese, S.: 3d-r2n2: A unified approach for single and multi-view 3d object reconstruction. In: {European Conference on Computer Vision (ECCV)}. pp. 628--644. Springer (2016)

\bibitem{cseke_tripathi_2025_pico}
Cseke, A., Tripathi, S., Dwivedi, S.K., Lakshmipathy, A., Chatterjee, A., Black, M.J., Tzionas, D.: {PICO}: Reconstructing {3D} people in contact with objects. In: {Computer Vision and Pattern Recognition (CVPR)}. pp. 1783--1794 (June 2025)

\bibitem{deitke2023objaverseXL}
Deitke, M., Liu, R., Wallingford, M., Ngo, H., Michel, O., Kusupati, A., Fan, A., Laforte, C., Voleti, V., Gadre, S.Y., VanderBilt, E., Kembhavi, A., Vondrick, C., Gkioxari, G., Ehsani, K., Schmidt, L., Farhadi, A.: {Objaverse-XL}: {A} universe of 10m+ {3D} objects. {arXiv}:2307.05663  (2023)

\bibitem{deitke2023objaverse}
Deitke, M., Schwenk, D., Salvador, J., Weihs, L., Michel, O., VanderBilt, E., Schmidt, L., Ehsani, K., Kembhavi, A., Farhadi, A.: Objaverse: {A} universe of annotated {3D} objects. In: {Computer Vision and Pattern Recognition (CVPR)}. pp. 13142--13153 (2023)

\bibitem{poco}
Dwivedi, S.K., Schmid, C., Yi, H., Black, M.J., Tzionas, D.: {POCO}: {3D} pose and shape estimation using confidence. In: {International Conference on {3D} Vision (3DV)}. pp. 85--95 (2024)

\bibitem{tokenhmr}
Dwivedi, S.K., Sun, Y., Patel, P., Feng, Y., Black, M.J.: {TokenHMR}: Advancing human mesh recovery with a tokenized pose representation. In: {Computer Vision and Pattern Recognition (CVPR)}. pp. 1323--1333 (2024)

\bibitem{psgn}
Fan, H., Su, H., Guibas, L.J.: A point set generation network for 3d object reconstruction from a single image. In: 2017 {IEEE} Conference on Computer Vision and Pattern Recognition, {CVPR} 2017, Honolulu, HI, USA, July 21-26, 2017. pp. 2463--2471. {IEEE} Computer Society (2017)

\bibitem{Fieraru_2021_AAAI}
Fieraru, M., Zanfir, M., Oneata, E., Popa, A.I., Olaru, V., Sminchisescu, C.: Learning complex {3D} human self-contact. In: {AAAI Conference on Artificial Intelligence}. pp. 1343--1351 (2021)

\bibitem{GemanMcclure}
Geman, S., Mcclure, D.E.: Statistical methods for tomographic image reconstruction. Bulletin of the International Statistical Institute  \textbf{52}(4),  5--21 (1987)

\bibitem{imgto3d_voxel1}
Girdhar, R., Fouhey, D., Rodriguez, M.D., Gupta, A.: Learning a predictable and generative vector representation for objects. In: {European Conference on Computer Vision (ECCV)}. vol.~9910, pp. 484--499 (2016)

\bibitem{goel2023humans}
Goel, S., Pavlakos, G., Rajasegaran, J., Kanazawa*, A., Malik*, J.: Humans in {4D}: Reconstructing and tracking humans with transformers. In: {International Conference on Computer Vision ({ICCV})}. pp. 14737--14748 (2023)

\bibitem{Hassan2021posa}
Hassan, M., Ghosh, P., Tesch, J., Tzionas, D., Black, M.J.: Populating {3D} scenes by learning human-scene interaction. In: {Computer Vision and Pattern Recognition (CVPR)}. pp. 14708--14718 (2021)

\bibitem{hasson2019obman}
Hasson, Y., Varol, G., Tzionas, D., Kalevatykh, I., Black, M.J., Laptev, I., Schmid, C.: Learning joint reconstruction of hands and manipulated objects. In: {Computer Vision and Pattern Recognition (CVPR)}. pp. 11807--11816 (2019)

\bibitem{lrm}
Hong, Y., Zhang, K., Gu, J., Bi, S., Zhou, Y., Liu, D., Liu, F., Sunkavalli, K., Bui, T., Tan, H.: {LRM}: {L}arge reconstruction model for single image to {3D}. {arXiv}: 2311.04400  (2023)

\bibitem{RICH}
Huang, C.H., Yi, H., Höschle, M., Safroshkin, M., Alexiadis, T., Polikovsky, S., Scharstein, D., Black, M.J.: Capturing and inferring dense full-body human-scene contact. In: {Computer Vision and Pattern Recognition (CVPR)}. pp. 13264--13275 (2022)

\bibitem{huang2022intercap}
Huang, Y., Taheri, O., Black, M.J., Tzionas, D.: {InterCap}: Joint markerless {3D} tracking of humans and objects in interaction from multi-view {RGB-D} images. {International Journal of Computer Vision (IJCV)}  \textbf{132}(7),  2551--2566 (2024)

\bibitem{stackflow}
Huo, C., Shi, Y., Ma, Y., Xu, L., Yu, J., Wang, J.: Stackflow: Monocular human-object reconstruction by stacked normalizing flow with offset. In: Proceedings of the Thirty-Second International Joint Conference on Artificial Intelligence, {IJCAI} 2023, 19th-25th August 2023, Macao, SAR, China. pp. 902--910. ijcai.org (2023)

\bibitem{real3d}
Jiang, H., Huang, Q., Pavlakos, G.: Real3d: Towards scaling large reconstruction models with real images. In: {International Conference on Computer Vision ({ICCV})}. pp. 5821--5833 (October 2025)

\bibitem{joo2021eft}
Joo, H., Neverova, N., Vedaldi, A.: Exemplar fine-tuning for {3D} human pose fitting towards in-the-wild {3D} human pose estimation. In: {International Conference on {3D} Vision (3DV)}. pp. 42--52 (2021)

\bibitem{kamakura1980grasp}
Kamakura, N., Matsuo, M., Ishii, H., Mitsuboshi, F., Miura, Y.: Patterns of static prehension in normal hands. American Journal of Occupational Therapy  \textbf{34}(7),  437--445 (1980)

\bibitem{gaussian_splat}
Kerbl, B., Kopanas, G., Leimk{\"{u}}hler, T., Drettakis, G.: 3d gaussian splatting for real-time radiance field rendering. {ACM} Trans. Graph.  \textbf{42}(4),  139:1--139:14 (2023)

\bibitem{Kocabas2021pare}
Kocabas, M., Huang, C.H.P., Hilliges, O., Black, M.J.: {PARE}: {P}art attention regressor for {3D} human body estimation. In: {International Conference on Computer Vision ({ICCV})}. pp. 11127--11137 (2021)

\bibitem{Kolotouros2019ConvolutionalMR}
Kolotouros, N., Pavlakos, G., Daniilidis, K.: Convolutional mesh regression for single-image human shape reconstruction. In: {Computer Vision and Pattern Recognition (CVPR)}. pp. 4496--4505 (2019)

\bibitem{unitepeople}
Lassner, C., Romero, J., Kiefel, M., Bogo, F., Black, M.J., Gehler, P.: Unite the people: {C}losing the loop between {3D} and {2D} human representations. In: {Computer Vision and Pattern Recognition (CVPR)} (2017)

\bibitem{li2021hybrik}
Li, J., Xu, C., Chen, Z., Bian, S., Yang, L., Lu, C.: {HybrIK}: {A} hybrid analytical-neural inverse kinematics solution for {3D} human pose and shape estimation. In: {Computer Vision and Pattern Recognition (CVPR)}. pp. 3383--3393 (2021)

\bibitem{li2022cliff}
Li, Z., Liu, J., Zhang, Z., Xu, S., Yan, Y.: {CLIFF}: {C}arrying location information in full frames into human pose and shape estimation. In: {European Conference on Computer Vision (ECCV)}. vol. 13665, pp. 590--606 (2022)

\bibitem{lin2023osx}
Lin, J., Zeng, A., Wang, H., Zhang, L., Li, Y.: One-stage {3D} whole-body mesh recovery with component aware transformer. In: {Computer Vision and Pattern Recognition (CVPR)}. pp. 21159--21168 (2023)

\bibitem{lin2021end-to-end}
Lin, K., Wang, L., Liu, Z.: End-to-end human pose and mesh reconstruction with transformers. In: {Computer Vision and Pattern Recognition (CVPR)}. pp. 1954--1963 (2021)

\bibitem{meshgraphormer}
Lin, K., Wang, L., Liu, Z.: Mesh graphormer. In: {International Conference on Computer Vision ({ICCV})} (2021)

\bibitem{zero1to3}
Liu, R., Wu, R., Hoorick, B.V., Tokmakov, P., Zakharov, S., Vondrick, C.: Zero-1-to-3: Zero-shot one image to {3D} object. {International Conference on Computer Vision ({ICCV})}  (2023)

\bibitem{liu2023grounding}
Liu, S., Zeng, Z., Ren, T., Li, F., Zhang, H., Yang, J., Li, C., Yang, J., Su, H., Zhu, J., et~al.: Grounding dino: Marrying dino with grounded pre-training for open-set object detection. arXiv preprint arXiv:2303.05499  (2023)

\bibitem{easyhoi}
Liu, Y., Long, X., Yang, Z., Liu, Y., Habermann, M., Theobalt, C., Ma, Y., Wang, W.: Easyhoi: Unleashing the power of large models for reconstructing hand-object interactions in the wild. In: {Computer Vision and Pattern Recognition (CVPR)}. pp. 7037--7047. Computer Vision Foundation / {IEEE} (2025)

\bibitem{SMPL:2015}
Loper, M., Mahmood, N., Romero, J., Pons-Moll, G., Black, M.J.: {SMPL}: {A} skinned multi-person linear model. {Transactions on Graphics (TOG)}  \textbf{34}(6),  248:1--248:16 (2015)

\bibitem{ma2025p3sam}
Ma, C., Li, Y., Yan, X., Xu, J., Yang, Y., Wang, C., Zhao, Z., Guo, Y., Chen, Z., Guo, C.: P3-sam: Native 3d part segmentation. arXiv preprint arXiv:2509.06784  (2025)

\bibitem{realfusion}
Melas{-}Kyriazi, L., Laina, I., Rupprecht, C., Vedaldi, A.: Realfusion 360{\textdegree} reconstruction of any object from a single image. In: {Computer Vision and Pattern Recognition (CVPR)}. pp. 8446--8455. {IEEE} (2023)

\bibitem{nerf}
Mildenhall, B., Srinivasan, P.P., Tancik, M., Barron, J.T., Ramamoorthi, R., Ng, R.: Nerf: Representing scenes as neural radiance fields for view synthesis. In: Vedaldi, A., Bischof, H., Brox, T., Frahm, J. (eds.) Computer Vision - {ECCV} 2020 - 16th European Conference, Glasgow, UK, August 23-28, 2020, Proceedings, Part {I}. Lecture Notes in Computer Science, vol. 12346, pp. 405--421. Springer (2020)

\bibitem{hand4whole}
Moon, G., Choi, H., Lee, K.M.: Accurate 3d hand pose estimation for whole-body 3d human mesh estimation. In: {Computer Vision and Pattern Recognition Workshops (CVPRw)}. pp. 2307--2316. {IEEE} (2022)

\bibitem{Mueller:CVPR:2021}
M{\"u}ller, L., Osman, A.A.A., Tang, S., Huang, C.H.P., Black, M.J.: On self-contact and human pose. In: {Computer Vision and Pattern Recognition (CVPR)}. pp. 9990--9999 (2021)

\bibitem{nam2024contho}
Nam, H., Jung, D.S., Moon, G., Lee, K.M.: Joint reconstruction of {3D} human and object via contact-based refinement transformer. In: {Computer Vision and Pattern Recognition (CVPR)} (2024)

\bibitem{deepsdf}
Park, J.J., Florence, P., Straub, J., Newcombe, R., Lovegrove, S.: Deepsdf: Learning continuous signed distance functions for shape representation. In: {Computer Vision and Pattern Recognition (CVPR)}. pp. 165--174 (2019)

\bibitem{SMPL-X:2019}
Pavlakos, G., Choutas, V., Ghorbani, N., Bolkart, T., Osman, A.A.A., Tzionas, D., Black, M.J.: Expressive body capture: {3D} hands, face, and body from a single image. In: {Computer Vision and Pattern Recognition (CVPR)}. pp. 10975--10985 (2019)

\bibitem{hamer}
Pavlakos, G., Shan, D., Radosavovic, I., Kanazawa, A., Fouhey, D., Malik, J.: Reconstructing hands in 3d with transformers. In: {Computer Vision and Pattern Recognition (CVPR)}. pp. 9826--9836. {IEEE} (2024)

\bibitem{tridi}
Petrov, I.A., Marin, R., Chibane, J., Pons-Moll, G.: Tridi: Trilateral diffusion of 3d humans, objects, and interactions. In: {International Conference on Computer Vision ({ICCV})}. pp. 5523--5535 (October 2025)

\bibitem{magic123}
Qian, G., Mai, J., Hamdi, A., Ren, J., Siarohin, A., Li, B., Lee, H., Skorokhodov, I., Wonka, P., Tulyakov, S., Ghanem, B.: Magic123: One image to high-quality 3d object generation using both 2d and 3d diffusion priors. In: The Twelfth International Conference on Learning Representations, {ICLR} 2024, Vienna, Austria, May 7-11, 2024. OpenReview.net (2024)

\bibitem{ravi2024sam2segmentimages}
Ravi, N., Gabeur, V., Hu, Y.T., Hu, R., Ryali, C., Ma, T., Khedr, H., Rädle, R., Rolland, C., Gustafson, L., Mintun, E., Pan, J., Alwala, K.V., Carion, N., Wu, C.Y., Girshick, R., Dollár, P., Feichtenhofer, C.: Sam 2: Segment anything in images and videos (2024)

\bibitem{rempe2021humor}
Rempe, D., Birdal, T., Hertzmann, A., Yang, J., Sridhar, S., Guibas, L.J.: {HuMoR}: {3D} human motion model for robust pose estimation. In: {International Conference on Computer Vision ({ICCV})}. pp. 11468--11479 (2021)

\bibitem{MANO:SIGGRAPHASIA:2017}
Romero, J., Tzionas, D., Black, M.J.: Embodied hands: Modeling and capturing hands and bodies together. {Transactions on Graphics (TOG)}  \textbf{36}(6) (Nov 2017)

\bibitem{rong2021frankmocap}
Rong, Y., Shiratori, T., Joo, H.: {FrankMocap}: {A} monocular {3D} whole-body pose estimation system via regression and integration. In: {International Conference on Computer Vision Workshops (ICCVw)}. pp. 1749--1759 (2021)

\bibitem{shimada2022hulc}
Shimada, S., Golyanik, V., Li, Z., P{\'e}rez, P., Xu, W., Theobalt, C.: {HULC}: {3D} human motion capture with pose manifold sampling and dense contact guidance. In: {European Conference on Computer Vision (ECCV)}. pp. 516--533 (2022)

\bibitem{Shimada2020PhysCapTOG}
Shimada, S., Golyanik, V., Xu, W., Theobalt, C.: {PhysCap}: {P}hysically plausible monocular {3D} motion capture in real time. {Transactions on Graphics (TOG)}  \textbf{39}(6),  235:1--235:16 (2020)

\bibitem{GRAB:2020}
Taheri, O., Ghorbani, N., Black, M.J., Tzionas, D.: {GRAB}: {A} dataset of whole-body human grasping of objects. In: {European Conference on Computer Vision (ECCV)} (2020)

\bibitem{lgm}
Tang, J., Chen, Z., Chen, X., Wang, T., Zeng, G., Liu, Z.: {LGM:} large multi-view gaussian model for high-resolution 3d content creation. In: Leonardis, A., Ricci, E., Roth, S., Russakovsky, O., Sattler, T., Varol, G. (eds.) Computer Vision - {ECCV} 2024 - 18th European Conference, Milan, Italy, September 29-October 4, 2024, Proceedings, Part {IV}. Lecture Notes in Computer Science, vol. 15062, pp. 1--18. Springer (2024)

\bibitem{makeit3d}
Tang, J., Wang, T., Zhang, B., Zhang, T., Yi, R., Ma, L., Chen, D.: Make-it-3d: High-fidelity 3d creation from {A} single image with diffusion prior. In: {Computer Vision and Pattern Recognition (CVPR)}. pp. 22762--22772. {IEEE} (2023)

\bibitem{sam3dteam2025sam3d3dfyimages}
Team, S.D., Chen, X., Chu, F.J., Gleize, P., Liang, K.J., Sax, A., Tang, H., Wang, W., Guo, M., Hardin, T., Li, X., Lin, A., Liu, J., Ma, Z., Sagar, A., Song, B., Wang, X., Yang, J., Zhang, B., Dollár, P., Gkioxari, G., Feiszli, M., Malik, J.: Sam 3d: 3dfy anything in images  (2025)

\bibitem{hunyuan3d22025tencent}
Team, T.H.: Hunyuan3d 2.0: Scaling diffusion models for high resolution textured 3d assets generation (2025)

\bibitem{triposr}
Tochilkin, D., Pankratz, D., Liu, Z., Huang, Z., Letts, A., Li, Y., Liang, D., Laforte, C., Jampani, V., Cao, Y.: Triposr: Fast 3d object reconstruction from a single image. CoRR  \textbf{abs/2403.02151} (2024)

\bibitem{tripathi2023deco}
Tripathi, S., Chatterjee, A., Passy, J.C., Yi, H., Tzionas, D., Black, M.J.: {DECO}: {D}ense estimation of {3D} human-scene contact in the wild. In: {International Conference on Computer Vision ({ICCV})}. pp. 8001--8013 (2023)

\bibitem{sv3d}
Voleti, V., Yao, C., Boss, M., Letts, A., Pankratz, D., Tochilkin, D., Laforte, C., Rombach, R., Jampani, V.: {SV3D:} novel multi-view synthesis and 3d generation from a single image using latent video diffusion. In: Leonardis, A., Ricci, E., Roth, S., Russakovsky, O., Sattler, T., Varol, G. (eds.) Computer Vision - {ECCV} 2024 - 18th European Conference, Milan, Italy, September 29-October 4, 2024, Proceedings, Part {I}. Lecture Notes in Computer Science, vol. 15059, pp. 439--457. Springer (2024)

\bibitem{pixel2mesh}
Wang, N., Zhang, Y., Li, Z., Fu, Y., Liu, W., Jiang, Y.: Pixel2mesh: Generating 3d mesh models from single {RGB} images. In: Ferrari, V., Hebert, M., Sminchisescu, C., Weiss, Y. (eds.) Computer Vision - {ECCV} 2018 - 15th European Conference, Munich, Germany, September 8-14, 2018, Proceedings, Part {XI}. Lecture Notes in Computer Science, vol. 11215, pp. 55--71. Springer (2018)

\bibitem{pflrm}
Wang, P., Tan, H., Bi, S., Xu, Y., Luan, F., Sunkavalli, K., Wang, W., Xu, Z., Zhang, K.: {PF-LRM:} pose-free large reconstruction model for joint pose and shape prediction. In: The Twelfth International Conference on Learning Representations, {ICLR} 2024, Vienna, Austria, May 7-11, 2024. OpenReview.net (2024)

\bibitem{moge}
Wang, R., Xu, S., Dai, C., Xiang, J., Deng, Y., Tong, X., Yang, J.: Moge: Unlocking accurate monocular geometry estimation for open-domain images with optimal training supervision. In: {Computer Vision and Pattern Recognition (CVPR)}. pp. 5261--5271. Computer Vision Foundation / {IEEE} (2025)

\bibitem{WangHoiTg}
Wang, Z., Zheng, Q., Ma, S., Ye, M., Zhan, Y., Li, D.: End-to-end {HOI} reconstruction transformer with graph-based encoding. In: {Computer Vision and Pattern Recognition (CVPR)}. pp. 27706--27715. Computer Vision Foundation / {IEEE} (2025)

\bibitem{wen2025reconstructing}
Wen, B., Huang, D., Zhang, Z., Zhou, J., Deng, J., Gong, J., Chen, Y., Ma, L., Li, Y.L.: Reconstructing in-the-wild open-vocabulary human-object interactions. In: {Computer Vision and Pattern Recognition (CVPR)}. pp. 17426--17436 (2025)

\bibitem{wu2024reconstructing}
Wu, J., Pavlakos, G., Gkioxari, G., Malik, J.: Reconstructing hand-held objects in 3d. arXiv preprint arXiv:2404.06507  (2024)

\bibitem{xie2023tprocigen}
Xie, X., Bhatnagar, B.L., Lenssen, J.E., Pons-Moll, G.: Template free reconstruction of human-object interaction with procedural interaction generation. In: {Computer Vision and Pattern Recognition (CVPR)} (2024)

\bibitem{xie2022chore}
Xie, X., Bhatnagar, B.L., Pons{-}Moll, G.: {CHORE:} c{o}ntact, human and object reconstruction from a single {RGB} image. In: {European Conference on Computer Vision (ECCV)}. vol. 13662, pp. 125--145 (2022)

\bibitem{xu2020ghum}
Xu, H., Bazavan, E.G., Zanfir, A., Freeman, W.T., Sukthankar, R., Sminchisescu, C.: {GHUM} {\&} {GHUML}: Generative {3D} human shape and articulated pose models. In: {Computer Vision and Pattern Recognition (CVPR)}. pp. 6183--6192 (2020)

\bibitem{instantmesh}
Xu, J., Cheng, W., Gao, Y., Wang, X., Gao, S., Shan, Y.: Instantmesh: Efficient 3d mesh generation from a single image with sparse-view large reconstruction models. CoRR  \textbf{abs/2404.07191} (2024)

\bibitem{vitpose}
Xu, Y., Zhang, J., Zhang, Q., Tao, D.: Vitpose: Simple vision transformer baselines for human pose estimation. In: Koyejo, S., Mohamed, S., Agarwal, A., Belgrave, D., Cho, K., Oh, A. (eds.) Advances in Neural Information Processing Systems 35: Annual Conference on Neural Information Processing Systems 2022, NeurIPS 2022, New Orleans, LA, USA, November 28 - December 9, 2022 (2022)

\bibitem{sam3dbody}
Yang, X., Kukreja, D., Pinkus, D., Fan, T., Park, J., Shin, S., Cao, J., Liu, J.W., Ugrinovic, N., Sagar, A., et~al.: Sam 3d body: Robust full-body human mesh recovery. In: {Computer Vision and Pattern Recognition (CVPR)}. pp. 7209--7219 (2026)

\bibitem{yang2023lemon}
Yang, Y., Zhai, W., Luo, H., Cao, Y., Zha, Z.J.: {LEMON}: {L}earning {3D} human-object interaction relation from {2D} images. In: {Computer Vision and Pattern Recognition (CVPR)} (2024)

\bibitem{zhang2020phosa}
Zhang, J.Y., Pepose, S., Joo, H., Ramanan, D., Malik, J., Kanazawa, A.: Perceiving {3D} human-object spatial arrangements from a single image in the wild. In: {European Conference on Computer Vision (ECCV)}. vol. 12357, pp. 34--51 (2020)

\bibitem{zhang2024telling}
Zhang, J., Herrmann, C., Hur, J., Chen, E., Jampani, V., Sun, D., Yang, M.H.: Telling left from right: Identifying geometry-aware semantic correspondence. In: {Computer Vision and Pattern Recognition (CVPR)} (2024)

\bibitem{zhang2023neuraldome}
Zhang, J., Luo, H., Yang, H., Xu, X., Wu, Q., Shi, Y., Yu, J., Xu, L., Wang, J.: Neuraldome: A neural modeling pipeline on multi-view human-object interactions. In: {Computer Vision and Pattern Recognition (CVPR)} (2023)

\bibitem{Zhang_2021_ICCV}
Zhang, S., Zhang, Y., Bogo, F., Pollefeys, M., Tang, S.: Learning motion priors for {4D} human body capture in {3D} scenes. In: {International Conference on Computer Vision ({ICCV})}. pp. 11343--11353 (2021)

\bibitem{zhao2024imhoi}
Zhao, C., Zhang, J., Du, J., Shan, Z., Wang, J., Yu, J., Wang, J., Xu, L.: I'm hoi: Inertia-aware monocular capture of 3d human-object interactions. In: {Computer Vision and Pattern Recognition (CVPR)}. pp. 729--741 (2024)

\end{thebibliography}

\clearpage

\begin{center}
  {\large\bfseries Supplementary Material for:\\[0.4em]
   \outTitleFULL \par}
  \vspace{1.2em}
  {Agniv Chatterjee\orcidlink{0000-0001-7199-1196} \quad
          Georgios Pavlakos\orcidlink{0000-0001-5821-1909} \par}
  \vspace{0.6em}
  {University of Texas at Austin \par}
  {\ttfamily agniv.chatterjee@utexas.edu, pavlakos@cs.utexas.edu \par}
\end{center}
\vspace{1.5em}

\titlerunning{\nameMethod}
\authorrunning{A.~Chatterjee and G.~Pavlakos}

\renewcommand{\thepage}{S.\arabic{page}}
\renewcommand{\thesection}{S.\arabic{section}}
\renewcommand{\thefigure}{S.\arabic{figure}}
\renewcommand{\thetable}{S.\arabic{table}}
\renewcommand{\theequation}{S.\arabic{equation}}
\setcounter{page}{1}
\setcounter{section}{0}
\setcounter{figure}{0}
\setcounter{table}{0}
\setcounter{equation}{0}

\noindent 
In this Supplementary Material, we provide implementation details, additional analysis, and extended results that were omitted from the main paper due to space constraints. 
We first expand on the \nameMethod pipeline in \cref{sec:supmat:impl_deets}, including details about multiview rendering of the \lrm mesh (which is used for keypoint detection, pointcloud segmentation and correspondence estimation), keypoint triangulation, object template alignment and pointcloud segmentation. 
We then present further qualitative results in \cref{sec:supmat:results}, including discussion (\cref{subsec:supmat:discuss}) and representative failure cases (\cref{subsec:supmat:fail}). 
\Cref{sec:supmat:contact_eval} provides contact evaluation details, and \cref{sec:supmat:interact_limit} discusses the interaction cues captured by our method as well as its limitations. 
In \cref{sec:supmat:experiments}, we provide additional quantitative results. 
Finally, \cref{sec:supmat:runtime} presents a runtime breakdown of the full pipeline for recovering human-object geometry from a single \rgb image.
\section{\nameMethod implementation details}
\label{sec:supmat:impl_deets}

\noindent
\textbf{Multi-view rendering:}
For keypoint detection, correspondence estimation, and mesh segmentation, we render the \lrm mesh from multiple viewpoints,
using various values for the elevation and azimuth angles.
Specifically, we pick angles at intervals of 30 degrees for the azimuth angle, while for the elevation we consider the angles of 0, $\pm30$, and $\pm60$ degrees.
This leads to a total count of 60 viewpoints.

\noindent
\textbf{\threeD keypoint estimation:}
For hand keypoint triangulation, hand detection may return multiple candidates for the same hand in one view.
We therefore retain a single candidate per side before triangulation.
If the corresponding body wrist (\ie, wrist keypoint in the body keypoints list) is available, we choose the candidate whose centroid is closest to the body wrist in the image plane:
\[
m^\star
=
\arg\min_m
\left\|
\mu\!\left(x_{h,s}^{(v,m)}\right) - x_{b,\mathrm{wrist}(s)}^{(v)}
\right\|_2,
\]
where $x_{h,s}^{(v,m)} \in \mathbb{R}^{21\times 2}$ is the $m$-th candidate and $\mu(\cdot)$ denotes the centroid over the 21 hand joints.
We use the centroid rather than the wrist keypoint of the candidate's hand because the wrist lies at the arm-hand boundary and is often poorly localized.
Thus, averaging over all 21 keypoints yields a more stable spatial summary of the detection bounding box.
If the body wrist is unavailable, we fall back to the candidate with the largest image-plane extent.
This produces at most one left-hand and one right-hand observation per view.

Next,  we provide details about the confidence filtering, triangulation, and aggregation, summarized in Sec. 3.3 of the main paper.

\noindent
\textit{Confidence filtering.}
For each keypoint index $k$, we keep only views whose confidence exceeds a threshold $\gamma$ (empirically set to 0.6):
\(
\mathcal{O}_k
=
\left\{
v \in \mathcal{V}
\;\middle|\;
c_k^{(v)} > \gamma
\right\}.
\)
If $|\mathcal{O}_k| < N_{\min} = 3$, the keypoint is marked as invalid and excluded from fitting. 

\noindent
\textit{Robust triangulation.}
For each keypoint, we triangulate candidate \threeD points using all pairs of observed views. We select the hypothesis with the largest consensus set, i.e., the most inliers under reprojection-error threshold \(\tau\) (set to 5.0). 
The final 3D point is obtained by nonlinear refinement over the inlier views ($\mathcal{I}_k^\star$):
\begin{equation}
X_k^\star
=
\arg\min_{X \in \mathbb{R}^3}
\sum_{v \in \mathcal{I}_k^\star}
\left\|
\pi_v(X)-x_k^{(v)}
\right\|_2^2.
\end{equation}

\noindent
\textit{Confidence aggregation and output.}
For each valid keypoint, we compute an aggregated reliability score as the mean confidence over its inlier views:
\(
\bar{c}_k
=
\frac{1}{|\mathcal{I}_k^\star|}
\sum_{v \in \mathcal{I}_k^\star} c_k^{(v)}.
\)
Applying this to body and hand joints yields
\(
X_b \in \mathbb{R}^{25\times 3},
\)
\(
X_{h,\mathrm{L}} \in \mathbb{R}^{21\times 3},
\)
and
\(
X_{h,\mathrm{R}} \in \mathbb{R}^{21\times 3},
\)
which we concatenate into
\(
X =
\left[
X_b;\,
X_{h,\mathrm{L}};\,
X_{h,\mathrm{R}}
\right]
\in \mathbb{R}^{67\times 3}
\)
and
\(
\bar{c} \in \mathbb{R}^{67}.
\)
The pair $(X,\bar{c})$ defines a sparse, confidence-weighted set of \threeD keypoints that are used in the next stage to fit \smplh to the \lrm mesh.

\noindent
\textbf{Object template alignment:}
For the template alignment, if reliable semantic correspondences between the template mesh and the segmented object point cloud cannot be established, we initialize alignment using centroid matching and scale normalization, and then refine the result using ICP only.
This typically occurs under severe occlusion of the object by the human or for very small objects, where the \lrm reconstruction does not recover the object well.

\noindent
\textbf{Pointcloud segmentation:}
We expand here on the segmentation procedure summarized in Sec. 3.5, in the main paper.
For each viewpoint $k$, we first compute a viewpoint quality score $Q_k$ that reflects both geometric coverage and segmentation reliability:
\begin{equation}
Q_k = \alpha \cdot \frac{|V_k^{\text{visible}}|}{|V|} + (1-\alpha) \cdot \frac{|M_k^{\text{human}}| + |M_k^{\text{object}}|}{|I_k|},
\end{equation}
where $V_k^{\text{visible}}$ are the vertices visible in view $k$, $V$ is the full vertex set, $M_k^{\text{human}}$ and $M_k^{\text{object}}$ are the human and object masks respectively, $I_k$ is the set of image pixels, and $\alpha \in [0,1]$ balances visibility and mask quality.
The viewpoint quality takes into account both the ratio of vertices visible in that viewpoint and the quality of the human and object masks. 

To better handle contact regions, we construct a multi-scale human-object boundary by dilating both masks with kernel sizes of 3, 5, and 7 and intersecting the dilated regions. 
We compute a boundary closeness weight using Euclidean distance transform from the boundary, normalized by a maximum distance threshold, and transformed through a non-linear function with an optional base score boost for highly overlapped regions.
This penalizes vertices near the boundary more heavily, while still allowing those farther away to be segmented.
The optional boost is applied when the size of the object is much smaller compared to the human - the boundary region covers most of the object.
For a projected vertex $p_i$, with distance $d(p_i, B)$ to this boundary $B$, we define a boundary-closeness weight:
\begin{equation}
    w_i^{\text{boundary}} = \left(\min\left(\frac{d(p_i, B)}{d_{\max}}, 1\right)\right)^\gamma + b,
\end{equation}
where $d_{\max}$ is a distance threshold for normalization, $\gamma\in[0.3, 0.8]$ is an adaptive exponent determined by the human-object overlap ratio, and $b \geq 0$ is an optional base boost for highly overlapped regions. This encourages accurate labeling near the human-object contact regions.
Finally, for each vertex $V_i$, we aggregate scores over all viewpoints $K_i$ where it is visible:
\begin{equation}
w_i = \sum_{k \in K_i} Q_k \cdot w_i^{\text{boundary}}(V_i),
\end{equation}
and apply a threshold 
on $\{w_i\}$ to obtain final human and object labels. 
Vertices labeled as non-human form the raw object point cloud. 
We then apply 
geometric filtering: statistical outlier removal, pruning vertices with too few neighbors within a local radius, and keeping only the largest DBSCAN cluster. 
The resulting clean object point cloud is composed with the fitted \smplh mesh.

\section{Further Qualitative Results}
\label{sec:supmat:results}

\begin{figure}[t]
    \centering
    \includegraphics[width=\textwidth]{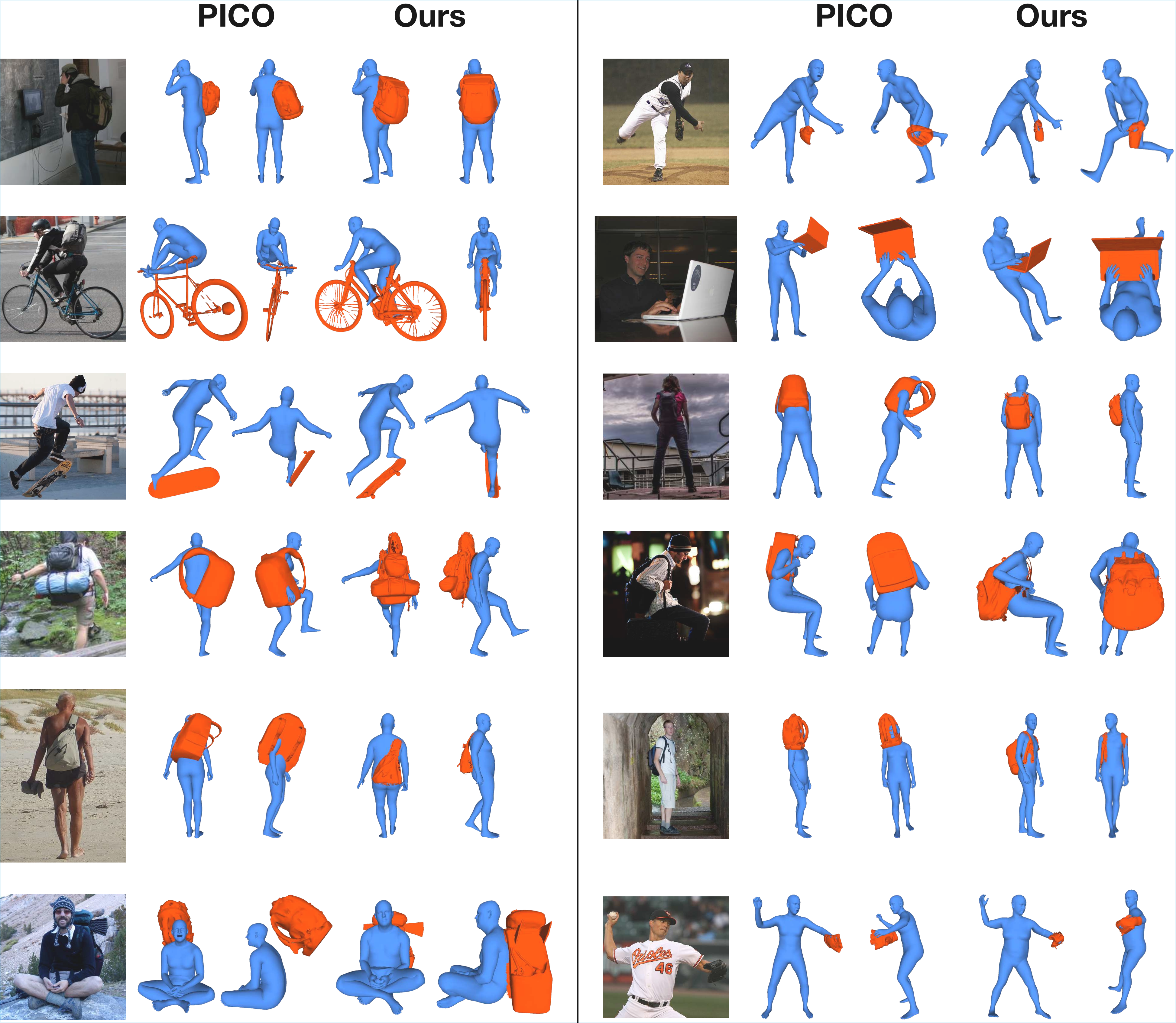}
    \caption{
    \textbf{Qualitative evaluation of PICO~\cite{cseke_tripathi_2025_pico} and \nameMethod on images from PICO-db~\cite{cseke_tripathi_2025_pico}.} 
    We visually compare the non-template version of \nameMethod with PICO, and
    output more accurate and physically plausible interactions, compared to PICO, without using ground-truth contact annotations.
    }
    \label{fig:supmat:pico_vis}
\end{figure}

\begin{figure}
    \centering
    \includegraphics[width=\textwidth]{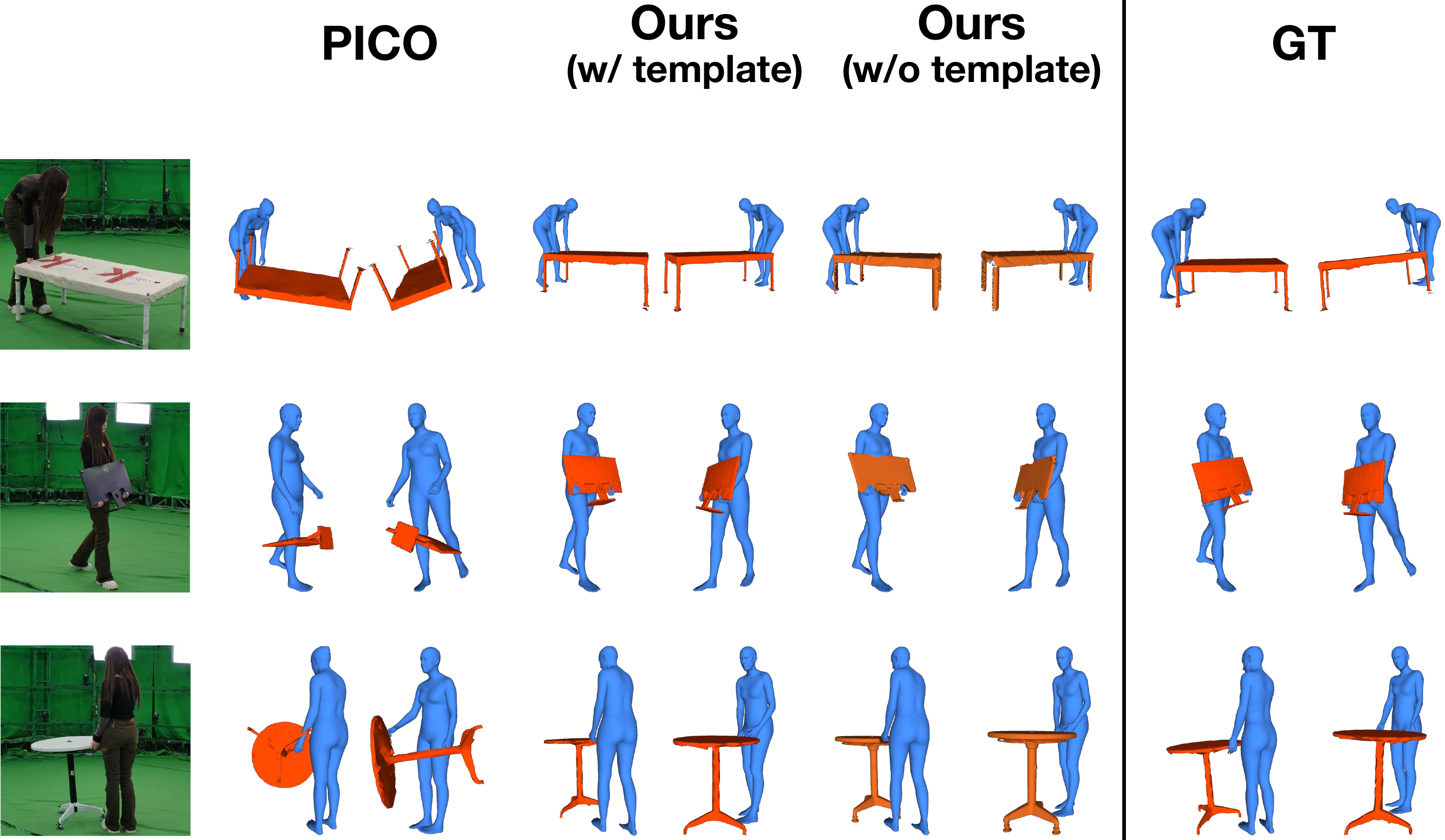}
    \caption{\textbf{Qualitative evaluation of PICO~\cite{cseke_tripathi_2025_pico} and \nameMethod on images from the HODome dataset~\cite{zhang2023neuraldome}.} 
    Our outputs, with or without templates, are more physically plausible and closer to the GT than PICO.}
    \label{fig:supmat:hodome_vis}
\end{figure}

\begin{figure}
    \centering
    \includegraphics[width=\textwidth]{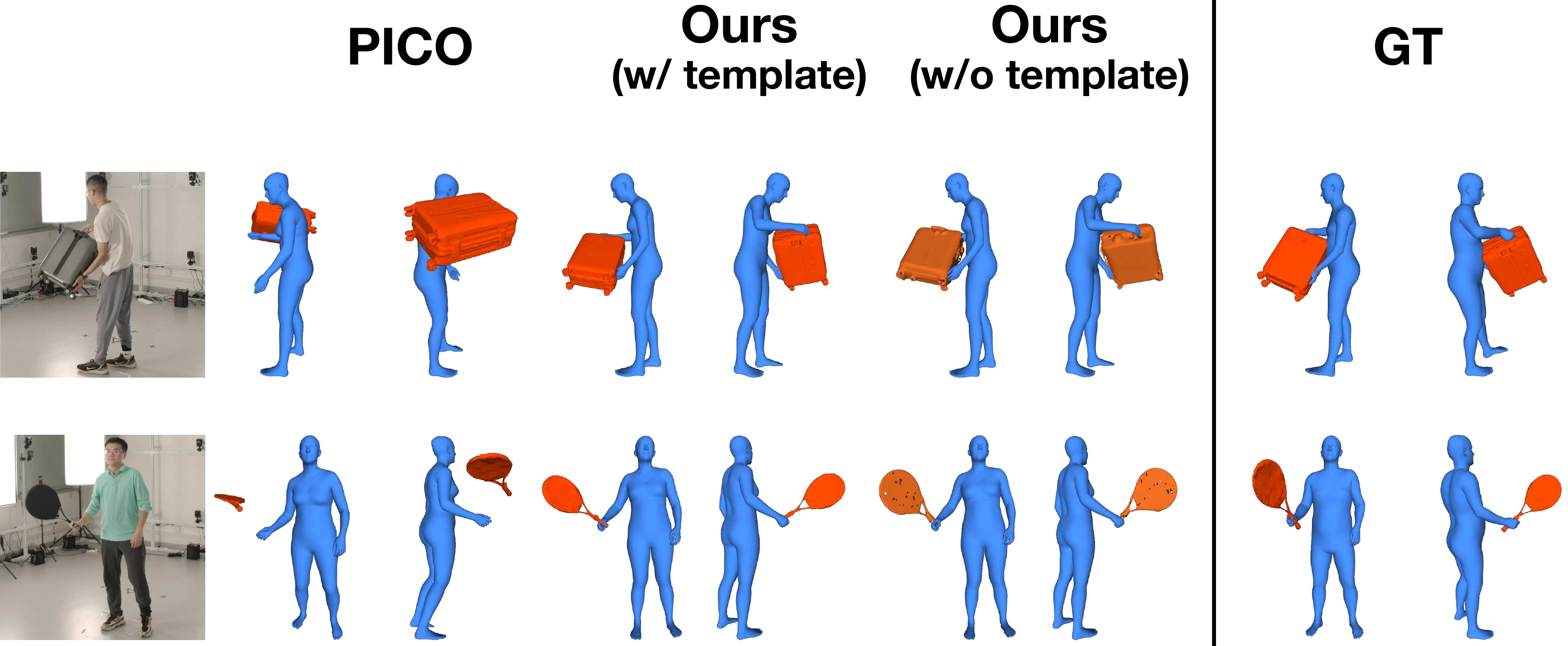}
    \caption{\textbf{Qualitative evaluation of PICO~\cite{cseke_tripathi_2025_pico} and \nameMethod on images from the IMHD dataset~\cite{zhao2024imhoi}.}}
    \label{fig:supmat:imhd_vis}
\end{figure}

We provide further qualitative results on PICO-db~\cite{cseke_tripathi_2025_pico} (\cref{fig:supmat:pico_vis}), HODome~\cite{zhang2023neuraldome} (\cref{fig:supmat:hodome_vis}) and IMHD~\cite{zhao2024imhoi} (\cref{fig:supmat:imhd_vis}) datasets. 
In these cases, we use PICO~\cite{cseke_tripathi_2025_pico} as a baseline for comparison.
For PICO-db, we visualize only the non-template variant of \nameMethod, since the dataset only provides retrieved object meshes, which may not be representative of the true object geometry.
Below, we discuss the qualitative trends observed across these datasets.

As can be seen from the qualitative results, our reconstructions better preserve the global object geometry, while accurately positioning it relative to the body. 
This is particularly visible for large carried or supported objects such as backpacks, suitcases, chairs, tables, and box-like objects, where PICO often produces implausible orientation or placement, whereas our method yields configurations that are more consistent with the image evidence.

On PICO-db (\cref{fig:supmat:pico_vis}), the non-template variant is visually more accurate than PICO in everyday in-the-wild scenes, despite the large variation in appearance, viewpoint, and object category. 
On HODome and IMHD (\cref{fig:supmat:hodome_vis} and
\cref{fig:supmat:imhd_vis} respectively), where template models are available, the template-based variant further improves geometric regularity and alignment. 
This is especially noticeable for rigid objects with canonical shapes, such as chairs, tables, and suitcases, where template alignment leverages a more precise object geometry while preserving the overall interaction. 
In contrast, the gap between the template and non-template variants is smaller for slender objects, such as bats, for which the main challenge is pose estimation rather than detailed surface recovery.

Finally, as can be seen from the visual results, even when the recovered object geometry is not very detailed, the reconstructed interaction is often more physically plausible than that of PICO, which is critical for downstream reasoning about contact, support, and affordances.

\subsection{Results Discussion}
\label{subsec:supmat:discuss}

For our quantitative evaluation on InterCap, we do not include HDM~\cite{xie2023tprocigen} in Tab. 1 of the main manuscript, since HDM is trained with ProciGen~\cite{xie2023tprocigen}, whose construction draws on both BEHAVE~\cite{bhatnagar2022behave} and InterCap~\cite{huang2022intercap}, so it has a competitive advantage over methods not trained on InterCap.

For the evaluation of PICO~\cite{cseke_tripathi_2025_pico} on HODome~\cite{zhang2023neuraldome} and IMHD~\cite{zhao2024imhoi}, we found that the retrieval-based strategy they proposed for the non-ground-truth-contact setting is often unsuccessful in these datasets.
In particular, several object categories in HODome, namely flower, box, pan, pillow, pink, and trashcan, and in IMHD, namely dumbbell, kettlebell, pan, and broom, do not have sufficiently similar counterparts in PICO-db, where the template retrieval is happening from. 
In addition, even when corresponding categories exist, differences in object geometry and interaction characteristics often affect retrieval quality. 
Accordingly, to obtain a more representative evaluation of PICO in this setting, we use the ground-truth template object meshes instead. 
To obtain paired human-object contact labels, we first use DECO~\cite{tripathi2023deco} to predict human vertices that are in contact.
From the predicted human vertices, we only consider the ones that are indeed in contact according to the ground-truth contact maps.
For these human vertices, we obtain the human-object vertex contact pairs from the ground truth human-object contact annotations.

Regarding our qualitative results, we use \lrm meshes as a geometric scaffold to pose the parametric human meshes without optimizing using any reprojection objectives.
Instead, we opt for \threeD consistency and accuracy of the interaction between humans and objects. 
Our observation is that the \lrm meshes generated by Hunyuan3D-2.0~\cite{hunyuan3d22025tencent} and other related approaches~\cite{real3d, triposr, sv3d, instantmesh} are generated in a coordinate frame different from the camera coordinate frame of the input image.
Thus, our reconstruction output would require an additional transformation to align with the input \rgb image.

\subsection{Failure Cases}
\label{subsec:supmat:fail}

\begin{figure}[t]
    \centering
    \includegraphics[width=\textwidth]{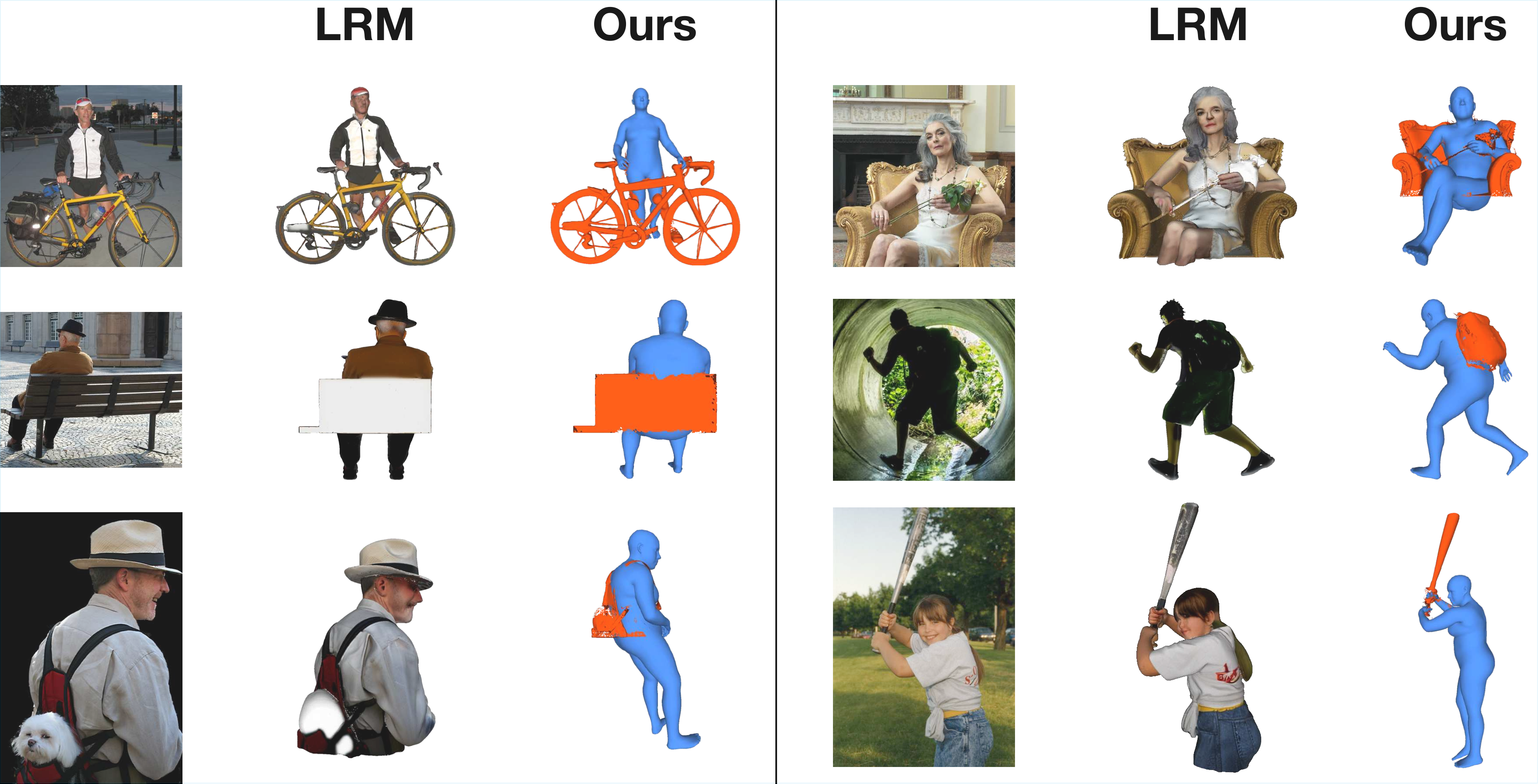}
    \caption{\textbf{Representative failure cases of \nameMethod}.
    Typical errors arise from incorrect \lrm reconstructions or imperfect point-cloud segmentation, leading to incomplete/warped geometry for large objects (mid-left), ambiguity when multiple objects fall into a single segmentation (top-right), objects being truncated in the image (top-right, bottom-left), leakage of non-object parts (e.g., shoes) into the object mask (top-left), and unfiltered scattered points around the object (mid-right, bottom row).
    }
    \label{fig:supmat:fails}
\end{figure}

We visualize some failures of \nameMethod in \cref{fig:supmat:fails}.
Failures typically occur either due to the quality of the \lrm reconstruction or due to the quality of the point cloud segmentation stage.
For larger, more complicated objects, the \lrm may fail to reconstruct the entirety of the object.
This also occurs in cases where the object is truncated.
Also, the \lrm might miss the boundary between the object and the background, causing elements of the background to seep into the reconstruction.
Finally, for instances where the human is interacting with multiple objects, the \lrm might reconstruct these additional objects.
Regarding the point cloud segmentation stage, this is reliant on the quality of the multi-view segmentation maps.
Thus, if the segmentation maps are noisy, it might be unable to filter out certain elements of the scene.

\begin{figure}
    \centering
    \includegraphics[width=\textwidth]{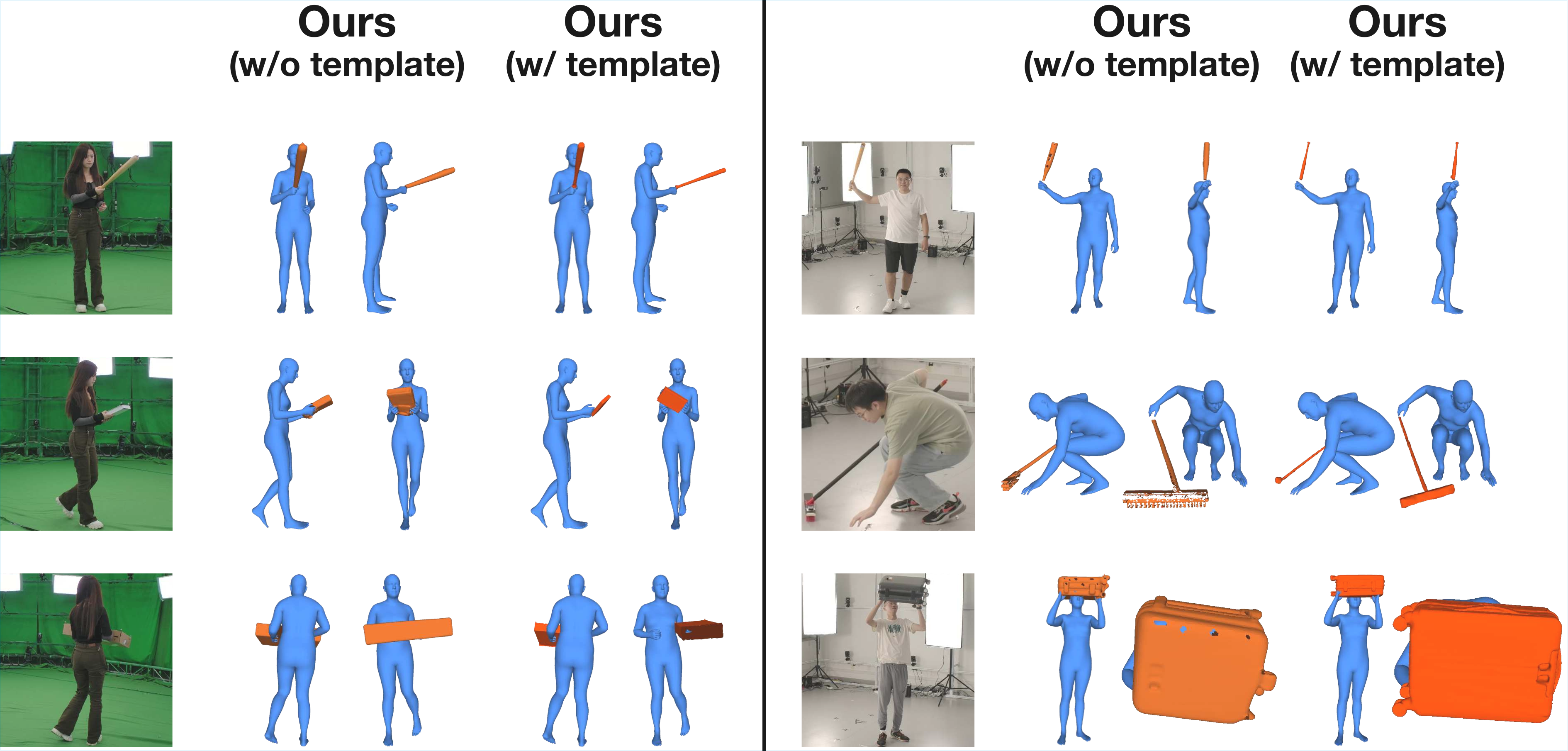}
    \caption{\textbf{Representative failure cases of the template alignment module}.
    Errors typically occur from incorrect semantic correspondences obtained due to the symmetrical nature of some objects, missing regions in the segmented object mesh, and occlusions, which cause correspondences only to visible regions.
    }
    \label{fig:supmat:fails_align}
\end{figure}

We also show representative failures of the template-alignment module in \cref{fig:supmat:fails_align}.
The quality of the recovered alignment is determined by the semantic correspondences established between template renders and \lrm renders.
This can be sensitive to appearance quality, lighting, occlusion, and object symmetry.
We observe two common failure modes. 
First, for nearly symmetric objects with less distinctive parts, the recovered alignment may be flipped. 
Second, when the object is only partially visible in the input image, the selected \lrm views tend to emphasize the visible portion, and the resulting correspondences may align the template only to that partial geometry rather than to the full object. 

Quantitatively, template alignment is most beneficial for thin or small objects, where the \lrm tends to produce incomplete geometry. For example, PA-CD$_o$ improves substantially after alignment for the bottle ($34.23 \rightarrow 9.97$ cm) and cup ($42.91 \rightarrow 18.64$ cm) categories.
For larger objects whose raw \lrm mesh is already reasonable, it can instead degrade the result, as observed for the trolley case ($17.46\rightarrow21.79$ cm), chair ($12.22\rightarrow18.77$ cm), and umbrella ($17.88\rightarrow26.76$ cm). 
The dominant error source is the symmetry-induced orientation flips noted above: mean rotation error reaches $90^\circ$-$120^\circ$ for symmetric objects, compared to $21.71^\circ$ for the chair, which provides stronger orientation cues. 
Scale recovery is generally reliable, with the umbrella being the main outlier (scale error 0.918, versus 0.27 on average for the remaining categories).
The box example in \cref{fig:supmat:fails_align} illustrates this effect.

\section{Contact Evaluation}
\label{sec:supmat:contact_eval}

For contact evaluation, we use the RICH checkpoint of DECO as our baseline.
This choice yields a fair comparison, as the best DECO checkpoint is obtained from training with InterCap in its training set. 
We do not evaluate contact metrics for the other \hoi reconstruction baselines because, except for HOI-TG, these methods explicitly rely on contact information as an input to reconstruction. 
By contrast, our method does not use contact annotations or contact priors at test time. 
We therefore compare against a state-of-the-art contact estimation network rather than against reconstruction methods that are given contact supervision.

To derive contact labels from our output, we first align the object template to the segmented object geometry. 
We then classify a human vertex as being in contact with the object if its Euclidean distance to the aligned object surface is smaller than 5 mm.

\section{Interaction Cues and Limitations}
\label{sec:supmat:interact_limit}

\nameMethod does not explicitly encode interaction using contact annotations, interaction labels, or object-specific priors. 
Instead, it leverages the geometric scaffold provided by the \lrm reconstruction, which implicitly captures relative arrangement and proximity cues between the human and the object. 
Based on this scaffold, we recover the human mesh, segment the object component, and optionally align an object template when one is available.

This formulation enables reconstruction of plausible human-object configurations \textit{without requiring explicit contact supervision}. 
As a result, the recovered geometry is also informative for downstream contact estimation, where our method performs competitively despite not predicting contact directly.

Our method can fail for small objects under severe occlusion.
For larger objects, partial occlusion is typically handled well via \lrm priors (bottom-left example, \cref{fig:supmat:pico_vis}; second example, \cref{fig:supmat:imhd_vis}), while occluded body regions are plausibly completed by \smplh. 
Purely \twoD-grounded baselines are equally, if not more, brittle under heavy occlusions. 
More broadly, without access to explicit object-shape information, the problem is particularly challenging.

\section{Additional Experiments}
\label{sec:supmat:experiments}

\begin{table}[t]
\centering
\begin{minipage}[t]{0.48\columnwidth}
\centering
\resizebox{\columnwidth}{!}{
\begin{tabular}{c|ccc}
\hline
\multirow{2}{*}{\begin{tabular}{c}Segmentation\\Used\end{tabular}} & PA-CD$_h$ & PA-CD$_o$ & PA-CD$_{h+o}$ \\
 & (cm) $\downarrow$ & (cm) $\downarrow$ & (cm) $\downarrow$ \\
\hline
 P3-SAM~\cite{ma2025p3sam} & 7.30 & 55.86 & 14.68 \\
 Ours & \bf 6.85 & \bf 20.74 & \bf 9.36 \\
\hline
\end{tabular}
}
\captionof{table}{\textbf{Comparison with P3-SAM.} Results presented on InterCap~\cite{huang2022intercap}.}
\label{tab:supmat:ablation_p3sam}
\end{minipage}
\hfill
\begin{minipage}[t]{0.48\columnwidth}
\centering
\resizebox{\columnwidth}{!}{
\begin{tabular}{c|ccc}
\hline
\multirow{2}{*}{\begin{tabular}{c}Method\\Used\end{tabular}} & PA-CD$_h$ & PA-CD$_o$ & PA-CD$_{h+o}$ \\
 & (cm) $\downarrow$ & (cm) $\downarrow$ & (cm) $\downarrow$ \\
\hline
 Kabsch & 7.16 & 19.98 & 7.53 \\
 Kabsch + ICP & \bf 6.96 & \bf 18.97 & \bf 7.45 \\
\hline
\end{tabular}
}
\captionof{table}{\textbf{Template alignment ablation.} Results presented on InterCap~\cite{huang2022intercap}.}
\label{tab:supmat:ablation_temp}
\end{minipage}
\end{table}

\subsection{Ablation Experiments}
\label{subsec:supmat:ablation}

Here, we provide additional ablation experiments.
First, we consider the use of alternative methods for point cloud segmentation.
Specifically, in \Cref{tab:supmat:ablation_p3sam}, we compare our proposed segmentation module with the recent P3-SAM method~\cite{ma2025p3sam}.
As can be seen, our strategy performs substantially better, 
which justifies the design of our approach.

Next, in \cref{tab:supmat:ablation_temp}, we ablate the stages of our template alignment pipeline. 
Using ICP on top of the initial weighted Kabsch fit consistently improves alignment across all metrics. 
The geometry-aware correspondences provide a reliable coarse similarity transform, while ICP further refines the local geometric fit to the segmented \lrm object point cloud.

Finally, we ablate the point cloud segmentation pipeline. 
Specifically, we analyze the effect of (i) visibility coverage and mask quality, which together define the viewpoint quality score $Q$, and (ii) the boundary-closeness weight $w^{\text{boundary}}$, which prevents confusion in vertices near human-object contact regions.
Dropping either visibility coverage or mask quality would mean the quality score is wholly computed from the value of the other term, while the boundary-closeness term can be dropped by setting it to unity.
The results in \cref{tab:supmat:ablation_pcseg} indicate that dropping any of these components leads to worse PA-CD.
This effectively highlights that all elements of our segmentation design are important for obtaining clean object point clouds from the combined \lrm mesh and, consequently, high-quality \hoi reconstructions.

\subsection{Individual object reconstruction}
\label{subsec:supmat:easyhoi}

Here, we provide more details about the EasyHOI-like experiment we present in the main manuscript.
The contact loss used in EasyHOI~\cite{easyhoi} is not directly applicable to our setting because it enforces contact based on hand-object overlap. 
In contrast, our notion of contact is not limited to the hands, and occlusion cannot be assumed to be a reliable indicator of contact.
Instead, we implement a contact-constrained variant that incorporates a distance-based contact term using ground-truth human contact vertices.
Concretely, we first extract the set of human vertices annotated as being in contact, and then penalize the distance from these vertices to the object. 
For each contact vertex, we compute the Euclidean distance to its nearest vertex on the transformed object mesh and define the penalty as the average of these minimum distances over all contact vertices.

\begin{table}[t]
\centering
\resizebox{0.8\columnwidth}{!}{
\begin{tabular}{ccc|ccc}
\hline
\multirow{2}{*}{\textbf{vis\_cover}} & \multirow{2}{*}{\textbf{mask\_quality}} & \multirow{2}{*}{\textbf{boundary\_dist}} & PA-CD$_h$ & PA-CD$_o$ & PA-CD$_{h+o}$ \\
 & & & (cm) $\downarrow$ & (cm) $\downarrow$ & (cm) $\downarrow$ \\ 
\hline
 \cmark & \cmark & \cmark & \bf 6.85 & \bf 20.74 & \bf 9.36 \\
 \xmark & \cmark & \cmark & 7.62 & 22.81 & 9.82 \\
 \cmark & \xmark & \cmark & 7.61 & 22.55 & 9.83 \\
 \cmark & \cmark & \xmark & 7.57 & 21.75 & 9.77 \\
\hline
\end{tabular}
}
\caption{\textbf{Ablation of point-cloud segmentation.}
Results are reported on InterCap~\cite{huang2022intercap}. 
The columns represent the visibility coverage (\textbf{vis\_cover}) and the mask quality (\textbf{mask\_quality}), together defining the viewpoint quality score $Q$, and the boundary-closeness weight (\textbf{boundary\_dist}).
Removing any component degrades performance, confirming each term’s contribution to clean point cloud segmentation.
}
\label{tab:supmat:ablation_pcseg}
\end{table}

\subsection{Open3DHOI baseline}
\label{subsec:supmat:open3dhoi}

In this subsection, we provide more details about our implementation of the Open3DHOI~\cite{wen2025reconstructing} baseline reported in Tab.~1 of the main paper. 
As the preprocessing code for Open3DHOI (object extraction and coarse alignment) has not been publicly released, we reimplement the coarse-reconstruction stage from the Open3DHOI manuscript, which reconstructs the human and object independently and places them in a common camera frame using monocular depth estimates.

We estimate per-frame \smplx parameters with OSX~\cite{lin2023osx} and obtain a watertight object mesh with InstantMesh~\cite{instantmesh}, the latter expressed in a normalized canonical frame with unknown scale, orientation, and translation. 
To recover the scene layout, we predict monocular depth with ZoeDepth~\cite{bhat2023zoedepth}. 
Since this depth is defined only up to a global scale, we calibrate it against the human: projecting the \smplx vertices into the image, we take the median ratio between camera-space and predicted depth over pixels within the person mask as the metric scale. 
Back-projecting the calibrated depth under the human and object masks yields a person and an object point cloud, with a median-absolute-deviation filter applied to the latter to suppress depth spikes at mask boundaries.

We align the human by translation only, shifting the \smplx root so that the mesh centroid coincides with the person point-cloud centroid, while keeping the OSX pose and scale fixed. 
For the object, scale and translation are taken from the corresponding InterCap ground-truth mesh (scale from the ratio of median radii, translation from the centroid), and orientation is recovered by ICP between the InstantMesh mesh and the object point cloud. 
The resulting human and object meshes, expressed in a shared camera frame, form the coarse initialization subsequently refined by the HOI-Gaussian optimizer.

\section{Runtime Analysis}
\label{sec:supmat:runtime}

All experiments are performed on an NVIDIA RTX 6000 Ada Generation GPU.
\Cref{tab:supmat:runtime} reports the runtime of each stage in our pipeline.
For the core pipeline, the total runtime is 344.40~s per image.

\begin{table}[!t]
    \centering
    \small
    \setlength{\tabcolsep}{6pt}
    \renewcommand{\arraystretch}{1.08}
    \begin{tabular}{lc}
        \toprule
        Stage & Time taken (s) \\
        \midrule
        \lrm reconstruction      & 189.00 \\
        Multi-view rendering        & 89.40 \\
        \twoD keypoint prediction     & 15.95 \\
        \threeD keypoint triangulation  & 7.95 \\
        Root fitting               & 10.70 \\
        Pose fitting               & 21.60 \\
        Mesh segmentation          & 9.80 \\
        \midrule
        \multicolumn{2}{l}{\textit{Optional template-alignment steps}} \\
        \addlinespace[2pt]
        Semantic correspondences   & 375.75 \\
        Template alignment         & 0.21 \\
        \bottomrule
    \end{tabular}
    \caption{\textbf{Runtime breakdown of each stage in our pipeline on a single NVIDIA RTX 6000 Ada GPU.}
    The first seven rows correspond to the core pipeline, whose total runtime is 344.40~s per image. 
    The last two rows report optional template-alignment steps, which are only used when template-based object alignment is enabled.}
    \label{tab:supmat:runtime}
\end{table}

We highlight that our core optimization stages require a total runtime of 32.3s, which is significantly lower compared to the three stages of our primary baseline PICO, which requires 475.84s (i.e., roughly half a minute for us, vs 8 minutes for PICO). 
The 2D keypoint prediction, 3D keypoint triangulation, root fitting, pose fitting, and mesh segmentation stages together require 66.0s, with 2D keypoint prediction (15.95s) and pose fitting (21.60s) the largest contributors among them.

The final two rows in Table~\ref{tab:supmat:runtime} correspond to the template-alignment steps that are optional and are only used when a template is available. 
In this setting, the additional cost is dominated by semantic correspondence estimation (375.75~s), while the final template alignment itself is negligible (0.21~s).
The runtime of semantic correspondence estimation is primarily incurred by running DINO, Stable Diffusion, and an aggregation network on all renders of the \lrm mesh and object template, and then performing dense vertex-level correspondence estimation.
Overall, these results show that the runtime of the core pipeline is primarily bottlenecked by the upstream \lrm reconstruction and rendering stages, whereas the optional template-based refinement is dominated by correspondence computation.

\end{document}